\documentclass[sigconf]{acmart}
\usepackage{amsmath} 
\usepackage{hyperref}
\usepackage[autostyle]{csquotes}
\usepackage{multirow}
\usepackage{array}
\usepackage[capitalize,noabbrev]{cleveref}
\usepackage{makecell}

\AtBeginDocument{%
  }

\copyrightyear{2025}
\acmYear{2025}
\setcopyright{acmlicensed}

\acmConference[KDD '25] {Proceedings of the 31st ACM SIGKDD Conference on Knowledge Discovery and Data Mining V.2}{August 3--7, 2025}{Toronto, ON, Canada.}
\acmBooktitle{Proceedings of the 31st ACM SIGKDD Conference on Knowledge Discovery and Data Mining V.2 (KDD '25), August 3--7, 2025, Toronto, ON, Canada}
\acmISBN{979-8-4007-1454-2/25/08}
\acmDOI{10.1145/3711896.3736835}

\begin{document}

\title{Advancing Tool-Augmented Large Language Models via Meta-Verification and Reflection Learning}


\author{Zhiyuan Ma}
\orcid{0009-0003-9188-7275}
\affiliation{%
  \institution{State Key Laboratory of Cognitive Intelligence, University of Science and Technology of China}
  \city{Hefei}
  \country{China}
}
\email{zhyma@mail.ustc.edu.cn}

\author{Jiayu Liu}
\affiliation{%
  \institution{State Key Laboratory of Cognitive Intelligence, University of Science and Technology of China}
  \city{Hefei}
  \country{China}
}
\email{jy251198@mail.ustc.edu.cn}

\author{Xianzhen Luo}
\affiliation{%
  \institution{Research Center for Social Computing and Interactive Robotics, Harbin Institute of Technology}
  \city{Harbin}
  \country{China}
}
\email{xzluo@ir.hit.edu.cn}

\author{Zhenya Huang}
\authornote{Zhenya Huang is the corresponding author.}
\affiliation{%
  \institution{State Key Laboratory of Cognitive Intelligence, University of Science and Technology of China}
  \city{Hefei}
  \country{China}
}
\email{huangzhy@ustc.edu.cn}

\author{Qingfu Zhu}
\affiliation{%
  \institution{Research Center for Social Computing and Interactive Robotics, Harbin Institute of Technology}
  \city{Harbin}
  \country{China}
}
\email{qfzhu@ir.hit.edu.cn}

\author{Wanxiang Che}
\affiliation{%
  \institution{Research Center for Social Computing and Interactive Robotics, Harbin Institute of Technology}
  \city{Harbin}
  \country{China}
}
\email{car@ir.hit.edu.cn}

\renewcommand{\shortauthors}{Ma et al.}

\begin{abstract}
Empowering large language models (LLMs) with effective tool utilization capabilities is crucial for enabling AI agents to solve complex problems. However, current models face two major limitations: (1) unreliable tool planning and invocation due to low-quality instruction datasets (e.g., widespread hallucinated API calls), and (2) weak tool reflection abilities (over 90\% of errors cannot be corrected) resulting from static imitation learning.
To address these critical limitations, we propose \textbf{Tool-MVR}, a novel Tool-Augmented LLM that achieves comprehensive System 2 reasoning through two key innovations. 
Specifically, we first introduce \textbf{M}ulti-\textbf{A}gent \textbf{M}eta-\textbf{V}erification (\textbf{MAMV}), a systematic pipeline that rigorously validates APIs, queries, and reasoning trajectories to construct ToolBench-V, a new high-quality instruction dataset that addresses the limitation of unreliable tool planning and invocation. Second, we propose \textbf{Explo}ration-based \textbf{Re}flection Learning (\textbf{EXPLORE}), which enhances tool reflection capabilities by leveraging tool feedback through a dynamic ``Error → Reflection → Correction'' learning paradigm, resulting in our reflection dataset ToolBench-R and addressing the critical weakness in tool reflection.
Finally, we obtain Tool-MVR by finetuning open-source LLMs (e.g., Qwen-7B) on both ToolBench-V and ToolBench-R.
Our experiments demonstrate that Tool-MVR achieves state-of-the-art performance on StableToolBench, surpassing both ToolLLM (by 23.9\%) and GPT-4 (by 15.3\%) while reducing API calls by 31.4\%, with strong generalization capabilities across unseen tools and scenarios. Additionally, on our proposed RefineToolBench, the first benchmark specifically designed to evaluate tool reflection capabilities. Tool-MVR achieves a 58.9\% error correction rate, significantly outperforming ToolLLM's 9.1\%.
\footnote{Codes and data are available at: \url{https://github.com/zhymma/Tool-MVR}}
\end{abstract}



\maketitle

\section{Introduction}
\label{sec:intro}
Large language models (LLMs) have shown impressive capabilities in various tasks, from natural language understanding to complex reasoning~\cite{zhao2023survey,ren2024survey,ding2024reasoning,fan2024survey, xue2024decompose, 10.1109/TKDE.2024.3400824,10506571,zhao2024comprehensive,liu2023guiding}. However, general-purpose LLMs face inherent limitations in accessing real-time data and specialized expertise~\cite{qu2024tool,wang2024tools,liu2024socraticlm,10.1145/3637528.3671841}. Similar to how humans leverage tools for problem-solving, the tool learning task enables LLMs to utilize external tools (e.g., weather forecast APIs, airline reservation APIs) for accessing real-time information and interacting with the physical world, expanding their capabilities~\cite{qin2024toollearningfoundationmodels}.

\begin{figure*}[t]
\centering
\includegraphics[width=\linewidth]{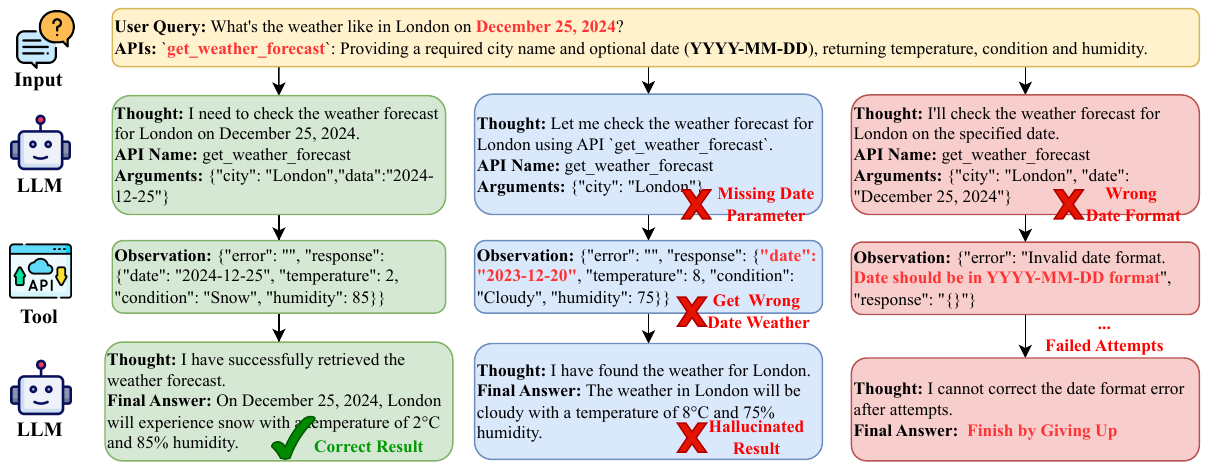}
\caption{Tool learning examples: given user queries and APIs, models generate API calls for answers. The green trajectory shows correct API calls, the blue trajectory shows hallucinated API calls, and the red trajectory demonstrates failed self-reflection.}
\label{fig1}
\end{figure*}

Tool learning requires multi-turn interactions between LLMs and external tools. As shown in the green trajectory of~\cref{fig1}, for a weather query ``What's the weather \ldots December 25, 2024?'', the model involves three key capabilities: (1) \textbf{Tool planning}: The LLM correctly identifies the need to use the weather forecast API. (2) \textbf{Tool invocation}: It then generates precise API calls with proper parameters (city=``London'', date=``2024-12-25''). (3) \textbf{Tool reflection}: It corrects errors through feedback when tool planning or invocation fails~\cite{hong2024advances,shi2025tool,plaat2024reasoning,ji2025test,Ma_Huang_Liu_Wang_Zhao_Li_2025,zhao2025unveiling}. Such a complex tool learning process requires System 2 reasoning - a deliberate, step-by-step approach that humans use for sophisticated problem-solving~\cite{kahneman2011thinking,evans2003two,stanovich2000individual,ijcai2024p725,tkde10946242}.

Recent studies have explored two main approaches to tackle tool learning task~\cite{qu2024tool,wang2024tools}. The first approach focuses on building Tool-Augmented agents using powerful closed-source LLMs through prompt engineering~\cite{DBLP:journals/tmlr/ChenM0C23,gao2023pal}, from ReAct's \enquote{thought-action-observation} format~\cite{yaoreact} to more sophisticated frameworks like ToolChain's A* search~\cite{zhuangtoolchain}. The second approach aims to enhance open-source LLMs' tool utilization capabilities through instruction tuning. ToolBench~\cite{qintoolllm}, currently the most widely-used dataset, leverages large-scale REST API collections and constructs extensive tool instruction data through depth-first search annotation. This dataset has enabled the development of various Tool-Augmented LLMs~\cite{chen2024advancing,yu2024steptool,DING2024110366}. For example, the instruction-tuned ToolLLM achieves performance comparable to ChatGPT~\cite{openai2022chatgpt}.

However, current tool learning research faces two critical limitations. First, mainstream tool instruction datasets (e.g., ToolBench) used for training Tool-Augmented LLMs suffer from severe quality issues. Recent analysis~\cite{iskander2024quality} reveals that in ToolBench~\cite{qintoolllm}, 57.3\% of queries contain unsolvable requests or incomplete information, and more critically, 74.4\% of API call trajectories exhibit hallucination behaviors (detailed in \cref{sec:3.2}). As illustrated by the blue trajectory in \cref{fig1}, which shows a typical hallucinated API call in ToolBench: (1) omitting the required date parameter and only calling the API with ``city: London''; and (2) incorrectly using current weather data (December 20) to answer a query about December 25. These poor-quality instructions severely impair the model's System 2 reasoning capabilities in both tool planning and tool invocation~\cite{li2024quantity}.

Furthermore, current Tool-Augmented LLMs lack tool reflection capabilities - a critical weakness is given that API call errors are inevitable in complex real-world scenarios~\cite{ji2023towards,renze2024self}. The ability to correct errors based on execution feedback is essential for robust tool utilization. However, we observe that ToolLLM abandons over 25\% of tasks because it falls into local traps when encountering errors, unable to correct wrong actions from multi-turn interactions. As shown in the red trajectory of \cref{fig1}, when encountering an incorrect date format error, the model fails to make the simple correction despite receiving explicit API feedback about the required YYYY-MM-DD format, and abandons the task. To further emphasize this phenomenon, we introduce RefineToolBench, the first benchmark designed to measure tool reflection capabilities (detailed in \cref{refinetoolbench}), which shows that ToolLLM successfully corrects errors in only 9.1\% of cases. This lack of adaptive tool reflection capability reveals that current Tool-Augmented LLMs have not yet achieved genuine System 2 reasoning~\cite{bașar2025well}.

To address these limitations, we develop a novel two-stage approach: First, unlike previous instruction construction methods that lack reliable verification mechanisms, we construct a new high-quality tool instruction dataset ToolBench-V through \textbf{M}ulti-\textbf{A}gent \textbf{M}eta-\textbf{V}erification (\textbf{MAMV}), a systematic pipeline that validates: (1) APIs and queries for tool planning - ensuring reliable API functionality and complete user requirements; and (2) API-call trajectories for tool invocation - guaranteeing accurate API call trajectories. This MAMV pipeline effectively addresses the data quality issues, establishing solid foundations for  tool planning and  tool invocation. Building upon this foundation, we introduce \textbf{Explor}ation-based \textbf{Re}flection Learning (EXPLORE) to enhance models' tool reflection capabilities. In contrast to previous training methods that focus solely on imitating successful expert trajectories while ignoring error cases~\cite{xiong2024watch,song2024trial}, EXPLORE actively explores error scenarios to identify model weaknesses and leverages execution feedback to construct reflection data. This forms a dynamic ``Error → Reflection → Correction'' learning paradigm, resulting in our reflection dataset ToolBench-R. Through this process, models learn to reflect and optimize wrong actions based on tool feedback, addressing the critical weakness in tool reflection capabilities.

Finally, by finetuning open-source LLMs (e.g., Qwen-2.5-7B~\cite{yang2024qwen2}) on both ToolBench-V and ToolBench-R, we obtain Tool-MVR that achieves sophisticated System 2 reasoning with deliberate tool planning, accurate tool invocation, and adaptive tool reflection. The main contributions of this work are as follows:
\begin{itemize}
    \item We propose Tool-MVR, achieving comprehensive System 2 reasoning capabilities with significantly improved performance, efficiency and generalization - surpassing GPT-4 by 15.3\% on StableToolBench while reducing API calls by 31.4\%, and maintaining strong performance across diverse unseen tools and scenarios.
    
    \item We develop MAMV for systematic instruction data verification, achieving substantial improvements in both query validity (from 52.7\% to 98.8\%) and trajectory accuracy (from 25.6\% to 81.3\%) in ToolBench-V.
    
    \item We introduce EXPLORE learning algorithm, enabling Tool-MVR to achieve a 58.9\% error correction rate on RefineToolBench, significantly outperforming ToolLLM's 9.1\%.
    
\end{itemize}

\section{Related Works}
\subsection{Tool Learning}
Tool learning has emerged as a crucial task for expanding LLMs' problem-solving abilities by enabling interaction with external tools to overcome inherent limitations such as outdated information and lack of real-time data access~\cite{zhao2023survey,ren2024survey}. Prior works have explored two main approaches. The first approach, focusing on prompt engineering with closed-source LLMs~\cite{shi2024learning,wu2024autogen,yuan2024easytool,DBLP:journals/tmlr/ChenM0C23,gao2023pal}, has evolved from ReAct's~\cite{yaoreact} basic thought-action format to more sophisticated frameworks. Chameleon~\cite{lu2024chameleon} enhanced compositional reasoning by orchestrating multiple tools with GPT-4, while ToolChain~\cite{zhuangtoolchain} improved efficiency through A* search in action space.
The second approach aims to enhance open-source LLMs through instruction tuning~\cite{DBLP:conf/iclr/GouSGSYHDC24,yuemammoth,chen2024towards,wu2024toolplanner,shi2024learning}. Toolformer~\cite{schick2023toolformer} pioneered this direction by embedding API calls into text generation. ToolBench~\cite{qintoolllm} made significant progress by establishing the largest tool learning dataset to date, encompassing 16,464 real-world APIs, and introduced ToolLLM with its depth-first search decision tree strategy (DFSDT) for trajectory annotation. This work provided a foundation for systematic tool learning research but was limited by dataset quality issues. Building upon ToolBench, TP-LLaMA~\cite{chen2024advancing} introduced preference learning to leverage failed explorations through DPO~\cite{rafailov2023direct,rafailov2024direct}. 

Most recently, works such as APIGen~\cite{liu2024apigen} and Tool-Ace~\cite{liu2025toolace} have introduced automated data generation frameworks aimed at improving tool utilization capabilities through methods like multi-stage API validation and diverse query styles. However, these approaches primarily focus on single-API instructions. In contrast, our work aims to enhance tool using and reflexion in complex multi-tool scenarios.

\subsection{Self-Reflection}
Recent work has explored feedback-based approaches for LLM self-reflection through various mechanisms~\cite{ji2023towards,renze2024self,zhao2024repair}. Self-Refine~\cite{madaan2024self} uses the same LLM for output generation and feedback-based improvements, while Reflexion~\cite{shinn2024reflexion} adds memory mechanisms to prevent recurring errors~\cite{pei2024memory}. For precise reasoning tasks, LeDex~\cite{jiang2024ledex} enhances code generation through explanation-based debugging, and CRITIC~\cite{goucritic} leverages external verification for mathematical reasoning. 
However, in tool learning, self-reflection remains limited. Existing approaches like DFSDT~\cite{qintoolllm} and AnyTool~\cite{duanytool} handle errors through simple backtracking or complete restarts, leading to inefficient reasoning. Our work advances feedback-driven reflection through an exploration-based learning algorithm, enabling direct error correction from tool feedback.

\begin{figure*}[t]
\centering
\includegraphics[width=\linewidth]{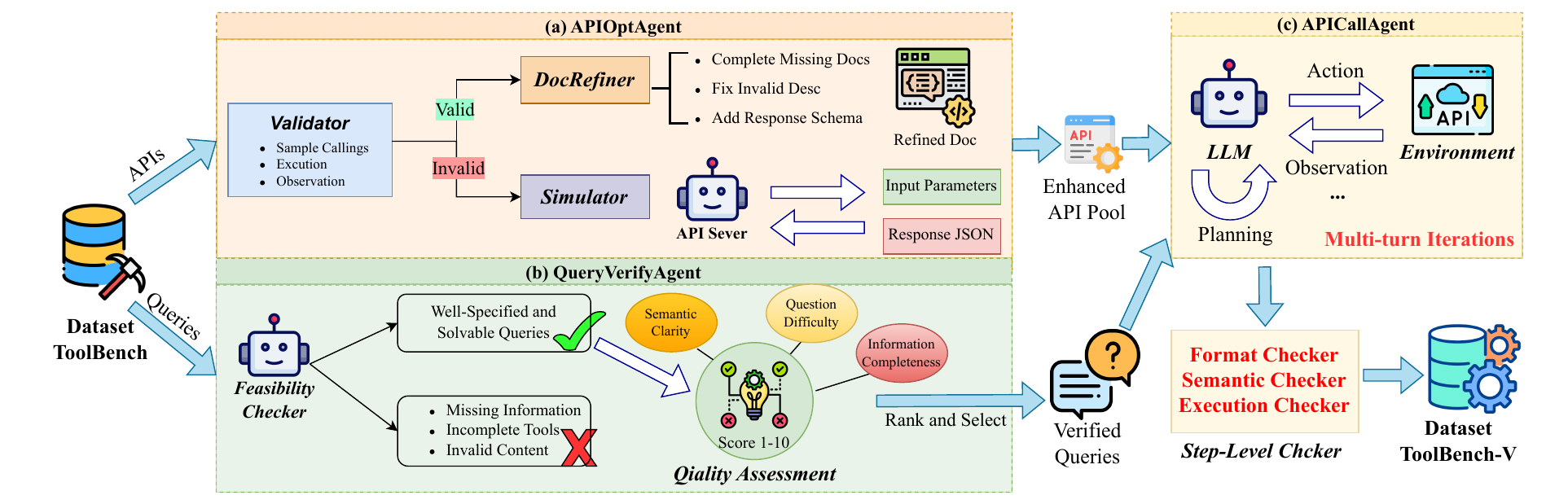}
\caption{Overview of Multi-Agent Meta-Verification (MAMV) framework. MAMV creates a new high-quality tool instruction dataset ToolBench-V via (a) APIOptAgent for API verification and optimization, (b) QueryVerifyAgent for query assessment and filtering, and (c) APICallAgent for step-level accurate API call trajectory generation.}
\label{fig: framework}
\end{figure*}

\section{Method}
In this section, we first provide a formal definition of tool learning task in~\cref{sec:3.1}. Then we present our approach to enhance Tool-Augmented LLMs through two key innovations: (1) developing MAMV to construct a new high-quality dataset ToolBench-V with 11,765 multi-turn tool interaction instances (\cref{sec:3.2}), and (2) developing EXPLORE, an exploration-based reflection learning algorithm that enables models to learn from tool feedback and correct errors in tool interaction (\cref{sec:3.3}).

\subsection{Preliminaries}
\label{sec:3.1}
The tool learning task requires models to effectively utilize external tools through multi-turn interactions~\cite{qu2024tool,wang2024tools}. As shown in \cref{fig1}, given a user query $Q$ (e.g., \enquote{What's the weather \ldots ?}) and a tool subset $T = \{t_1, \ldots, t_n\}$ (e.g., weather forecast APIs), the agent's objective is to utilize appropriate tools to acquire information and derive an answer. Let $\mathcal{T}$ denote the tool learning process:
\begin{equation}
    \mathcal{T}: Q \times T \rightarrow (S, r),
\end{equation}
where $S$ represents the reasoning trajectory and $r$ the final answer. The model gradually solves problems through multiple rounds of interaction. Each round $i$ consists of: (1) action $a_i = (t_i, p_i)$: selecting tool $t_i \in T$ and generating calling parameters $p_i$ (e.g., selecting weather API and specifying parameters like city=``London'', date=``2024-12-25''); (2) observation $o_i = t_i(p_i)$: executing the call and obtaining API feedback information $o_i$ (e.g., receiving weather forecast data or error messages about incorrect date format). Over the course of interaction, the model forms a reasoning trajectory and generates the final answer:
\begin{equation}
    (S, r) = ((a_1, o_1, \ldots, a_t, o_t), r).
\end{equation}
Based on the API responses in trajectory $S$, the model synthesizes a final answer $r$ (e.g., \enquote{The weather will be \ldots}). The effectiveness of solution is evaluated by determining whether trajectory $S$ and answer $r$ adequately address query $Q$ (details in \cref{sec:benandval}).

Unlike System 1's fast, intuitive thinking process, tool learning requires sophisticated System 2 reasoning - a deliberate, step-by-step problem solving approach~\cite{evans2003two,hong2024advances,plaat2024reasoning}. Specifically, this involves three key capabilities: (1) tool planning: analyzing query requirements and available APIs to determine the necessary tools; (2) tool invocation: constructing API calls with appropriate parameters (e.g., date=``2024-12-25''); and (3) tool reflection: adjusting tool planning and invocation based on execution feedback (e.g., modifying date format or selecting alternative APIs). In this work, we aim to enhance open-source LLMs' System 2 reasoning capabilities through these three aspects, enabling them to achieve the tool utilization capability of powerful closed-source models like GPT-4~\cite{achiam2023gpt4}.

\subsection{Multi-Agent Meta-Verification Pipeline}
\label{sec:3.2}
While instruction quality is crucial for model performance in specialized domains, existing mainstream tool instruction datasets primarily focus on quantity over quality~\cite{liu2024coachlm,li2024quantity}. Recent analysis from Quality Matters~\cite{iskander2024quality} reveals systematic issues in current datasets' construction methodology. These issues arise primarily from the lack of systematic verification in data collection process. Specifically, ToolBench~\cite{qintoolllm} relies heavily on ChatGPT~\cite{openai2022chatgpt} for data generation and employs a depth-first search strategy for trajectory annotation, without incorporating reliable verification mechanisms. This leads to the propagation of model hallucinations and annotator biases into the instruction data.

Specifically, we first analyze the representative dataset ToolBench~\cite{qintoolllm}, following the evaluation protocol in Quality Matters~\cite{iskander2024quality}. As shown in Table~\ref{tab:metrics}, our analysis reveals significant quality issues: For query quality, approximately 47.3\% of queries exhibit three critical issues: (1) information completeness: queries frequently omitting essential parameters (e.g., \enquote{Recommend romantic venues} missing location specification); (2) logical consistency: requirements often containing contradictions (e.g., booking past dates); and (3) API feasibility: queries requesting operations beyond API capabilities.
For API-call quality, about 74.4\% of trajectories contain three systematic problems: (1) parameter accuracy: 47.2\% of API calls containing incorrect formats or values; (2) solution completeness: 33.1\% of trajectories missing critical API calls; and (3) trajectory minimality: redundant API calls appearing as repeated queries. 
Such hallucination-prone instruction datasets like ToolBench lead to ToolLLM learning incorrect tool reasoning patterns, fundamentally limiting its tool planning and invocation capabilities.

To address the severe quality issues in existing tool learning datasets, we propose \textbf{M}ulti-\textbf{A}gent \textbf{M}eta-\textbf{V}erification (\textbf{MAMV}), a novel pipeline that systematically validates three meta-components in tool learning: APIs, user queries, and API call trajectories. Through collaborative verification, MAMV performs: (1) API verification (for tool planning - ensuring reliable API functionality and clear documentation), (2) query verification (for tool planning - ensuring query completeness and feasibility for tool selection), and (3) trajectory verification (for tool invocation - ensuring accurate parameter generation and efficient execution flow). Using MAMV, we improve ToolBench to ToolBench-V. Formally, each instance in ToolBench-V can be represented as:
\begin{equation}
    V = \{(Q, T, S)\},
\end{equation}
where $Q$ represents the verified queries from ToolBench, $T$ the enhanced API pool, and $S$ the newly constructed API call trajectory with step-wise verification.
MAMV employs multiple agents powered by GPT-4. Each agent focuses on a specific verification task, collaboratively ensuring comprehensive quality control.

\begin{table}[t]
\centering
\caption{Evaluation of instruction dataset quality scores (\%).}
\label{tab:metrics}
\small
\begin{tabular}{l|ccc|c}
\toprule
\textbf{Query Quality} & \textbf{Spec.} & \textbf{Coh.} & \textbf{Solv.} & \textbf{Overall} \\
\midrule
ToolBench & 79.6 & 77.9 & 81.8 & 52.7 \\
\textbf{ToolBench-V} & \textbf{99.9} & \textbf{99.0} & \textbf{99.9} & \textbf{98.8} \\
\midrule
\textbf{API-Call Quality} & \textbf{Align.} & \textbf{Suff.} & \textbf{Min.} & \textbf{Overall} \\
\midrule
ToolBench & 52.1 & 66.4 & 54.9 & 25.6 \\
\textbf{ToolBench-V} & \textbf{90.6} & \textbf{96.5} & \textbf{81.9} & \textbf{81.3} \\
\bottomrule
\end{tabular}
\end{table}

\subsubsection{API Verification}
We first focus on verifying and enhancing both API functionality and documentation. Leveraging RapidAPI Hub~\cite{rapidapi2024}, the world's largest API marketplace, we selected over 12,000 real-world APIs spanning 49 core domains (e.g., social media, travel services). These professionally maintained APIs support diverse practical applications, denoted as $T_{all} = \{t_1, \ldots, t_n\}$. However, these third-party APIs often present two major challenges: service instability and inconsistent documentation quality with incomplete parameter descriptions.
To address these challenges, we develop \textbf{APIOptAgent} as part of our MAMV pipeline (Figure~\ref{fig: framework}(a)). Specifically, we first employ GPT-4 to generate and execute diverse request samples for feedback collection. The \textbf{Validator} then assesses API validity based on these results, identifying common failure cases (e.g., ``API doesn't exist'', ``ACCESS\_DENIED''). For invalid APIs, our \textbf{Simulator} serves as a virtual API server, responding to requests with semantically appropriate JSON responses containing simulated data derived from API documentation to ensure continuous API functionality. For valid APIs, \textbf{DocRefiner} enhances documentation by adding missing parameter specifications (types, defaults, ranges) and standardizing descriptions based on API response analysis.
Through this process, we transformed 11,302 original APIs into an optimized pool: validating 6,770 functional APIs, enhancing 6,334 API documents, and providing simulation alternatives for 4,500+ unavailable APIs. This verification ensures both API functionality and clear documentation for effective tool planning (details in~\cref{sec:apiopt}).

\subsubsection{Query Verification}
User queries are crucial for accurate user intent analysis and appropriate tool selection. In ToolBench~\cite{qintoolllm}, approximately 80,000 initial queries were generated by ChatGPT~\cite{openai2022chatgpt}, covering various API usage scenarios. For each query $Q$, we select a relevant subset $T \subseteq T_{all}$ to address it. However, these ChatGPT-generated queries often contain unsolvable cases that hinder proper requirement analysis and tool mapping. For instance, queries like \enquote{I'm interested in a famous actor's career. Could you provide detailed information about their filmography?} lack essential information (actor name specification), making it impossible to map to appropriate APIs and parameters. Such incomplete queries lead to incorrect tool selection and flawed reasoning patterns, fundamentally limiting the model's tool planning capabilities.
We tackle these limitations by developing \textbf{QueryVerifyAgent} with two key components, as illustrated in~\cref{fig: framework}(b). The \textbf{Feasibility Checker} utilizes GPT-4 to filter out unsolvable queries due to missing information, logical contradictions, or lack of necessary tools. The \textbf{Quality Assessor} then evaluates remaining queries across three dimensions: semantic clarity, information completeness, and reasoning complexity, using a 1-10 scoring scale. We retain only high-quality queries (scores 8-10) to ensure query solvability, complexity, and completeness (see \cref{sec:queryverify} in Appendix). This rigorous filtering process yielded 23,903 high-quality queries from an initial pool of 62,147 (detailed implementation in~\cref{sec:queryverify}).

\subsubsection{Trajectory Verification}
\label{sec:trajectory_verification}
The quality of API call trajectories directly impacts models' tool invocation capabilities. The previous DFSDT approach, which combines ChatGPT with DFS algorithm for API sequence search, results in over 47\% parameter errors, 33\% incomplete solutions, and 45.1\% redundant trajectories. By focusing solely on final answer accuracy while ignoring intermediate reasoning processes, DFSDT generates numerous intermediate errors and hallucinated behaviors.

These challenges stem from the absence of verification in existing methods. To overcome these limitations, we develop \textbf{APICallAgent}, a multi-turn interactive framework with step-wise verification mechanism. As shown in \cref{fig: framework}(c), APICallAgent operates through multi-round interactions among user queries, LLM reasoning (GPT-4), and API environment feedback. The agent follows a systematic cycle of planning (chain-of-thought reasoning for tool selection), action (parameter generation), and observation (execution feedback processing) to ensure logical consistency throughout the reasoning process. Inspired by CodeAct~\cite{wangexecutable}, we adopt Python function calls instead of JSON format to better leverage LLMs' inherent coding capabilities, thereby improving API calling accuracy.

To ensure reliable API call trajectories, we designed a step-wise verification mechanism with three components: (1) \textbf{Format Checker} ensures final answer generation through syntax parsing, (2) \textbf{Semantic Checker} validates comprehensive problem resolution through constraint verification, and (3) \textbf{Execution Checker} examines each execution round to ensure positive contribution while eliminating redundant calls and hallucinated behaviors. This comprehensive verification ensures high-quality API call trajectories (detailed implementation in~\cref{sec:apicall}).

\subsubsection{ToolBench-V}
Hence, our MAMV pipeline creates ToolBench-V, a high-quality dataset of 11,765 multi-turn tool interaction instances. Through systematic verification, we effectively address the quality issues in existing datasets - achieving significant improvements in both query validity (from 52.7\% to 98.8\%) and trajectory accuracy (from 25.6\% to 81.3\%). The well-verified queries enhance models' ability to understand user intent and select appropriate tools, while accurate trajectories guide proper API parameter generation. This comprehensive meta-verification approach establishes a solid foundation for System 2 reasoning by ensuring reliable tool planning and efficient tool invocation capabilities.

\subsection{Exploration-based Reflection Learning}
\label{sec:3.3}

Currently, existing Tool-Augmented LLMs rely solely on static imitation learning with correct expert trajectories, inevitably leading to brittle models that struggle with error recovery~\cite{xiong2024watch,song2024trial}. Our analysis reveals that ToolLLM abandons 25\% of tasks when encountering execution failures, failing to recover and continue reasoning. To systematically verify this limitation, we introduce RefineToolBench, the first benchmark for evaluating models' tool reflection capabilities, where ToolLLM achieves only a 9.1\% error correction rate (details in~\cref{sec:4.5}). This poor tool reflection ability significantly limits models' practical applications in complex scenarios.

\subsubsection{EXPLORE}
\label{sec:explore}

\begin{figure}[t]
\centering
\includegraphics[width=\linewidth]{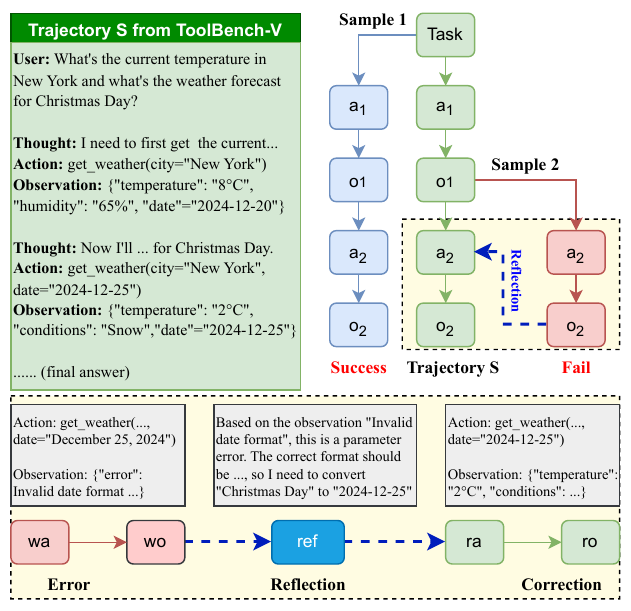}
\caption{Exploration-based Reflection Learning (EXPLORE): performs step-level exploration with sampling, enabling ``Error→Reflection→Correction'' learning paradigm.}
\label{fig3}
\end{figure}

In response to these limitations, we introduce EXPLORE, an exploration-based reflection learning algorithm enabling systematic error correction through a dynamic ``Error → Reflection → Correction'' learning paradigm. As illustrated in~\cref{fig3}, EXPLORE actively samples step-level error actions online to identify model weaknesses and leverages tool feedback to construct reflection data using GPT-4.
We categorize tool learning errors into two types: (1) API Calling errors (execution failures due to missing parameters, type mismatches, or incorrect API names); and (2) API Planning errors (successful but task-irrelevant executions due to incorrect tool selection or parameter content). Unlike previous approaches that discard failed cases~\cite{qintoolllm}, EXPLORE utilizes these errors as valuable learning signals, enabling models to reflect on mistakes and improve multi-turn interaction capabilities.

Building upon the foundational capabilities from the first stage, EXPLORE uses verified ToolBench-V trajectories as references for self-improvement. For a trajectory $S = (a_1, o_1, \ldots, a_n, o_n)$, we randomly select steps $o_t$ as exploration points. As shown in \cref{fig3}, for a weather query, the verified trajectory $S$ demonstrates correct API usage. Through multiple sampling, we construct an exploration tree where successful explorations (blue) lead to task completion, while failed attempts (red) trigger reflection generation. For failed attempts, the exploration process with $o_t$ as the starting point can be formalized as:
\begin{equation}
    E_t = \{(\text{wa}, \text{wo}, \text{ra}) |  \text{ra} = a_{t+1}\},
\end{equation}
where $\text{wa}$ represents wrong actions, $\text{wo}$ represents the feedback, and $\text{ra} = a_{t+1}$ the corresponding correct action. This reflection process is crucial, enabling models to learn from feedback and adjust tool planning or invocation.

\subsubsection{ToolBench-R}
\label{sec:toolbench-r}

Through this dynamic exploration process, we construct ToolBench-R, a reflection-oriented dataset:
\begin{align}
    R = \{&(Q, T, S_{<t}, \text{wa}, \text{wo}, (\text{ref}, \text{ra})) \mid \nonumber \\
    &(\text{wa}, \text{wo}, \text{ra}) \in Et\},
\end{align}
where each instance captures a complete reflection process with context: the query $Q$, tools $T$, and historical interactions $S_{<t}$.

ToolBench-R is constructed by leveraging GPT-4 to generate structured reflection content for each error case. This involves analyzing execution feedback for issues, developing correction strategies, and formulating precise execution plans. For example, when encountering date format errors in weather queries, GPT-4 analyzes the tool feedback to generate structured reflections that guide parameter correction from ``December 25, 2024" to ``2024-12-25". This structured reflection content forms the basis of ToolBench-R.

The reflection process emphasizes adaptive tool reflection through active processing and learning from real-time feedback. Each instance in ToolBench-R includes analytical (error analysis, correction strategy) and practical (concrete execution plans) components, enabling comprehensive tool reflection capabilities. This systematic approach results in ToolBench-R having 3,625 reflection instances covering diverse error types, robustly handling errors in complex real-world scenarios (further details in Appendix~\cref{sec:reflection}).

\subsection{Training}
We train our model through Supervised Fine-Tuning (SFT) to develop three System 2 reasoning capabilities. The training objective consists of two components corresponding to tool planning, invocation, and reflection abilities.

For ToolBench-V, which focuses on tool planning and invocation capabilities, we optimize the action prediction loss at each interaction step $t$ in sequence $S$:
\begin{equation}
    \mathcal{L}_V = -\sum_{x\in V}\sum_{t=1}^{|S|} \log p(a_t|Q, T, a_{<t}, o_{<t}),
\end{equation}
where $x=(Q,T,S)$ denotes instances from the meta-verified dataset.

For ToolBench-R, which enhances tool reflection capabilities through reflection learning, we optimize the joint prediction of reflection and corrected action:
\begin{equation}
    \mathcal{L}_R = -\sum_{y\in R} \log p((\text{ref},\text{ra})|Q,T,S_{<t},\text{wa},\text{wo}),
\end{equation}
where $y=(Q,T,S_{<t},\text{wa},\text{wo},(\text{ref},\text{ra}))$ denotes instances from the reflection dataset.
The final training objective combines both losses:
\begin{equation}
    \mathcal{L} = \mathcal{L}_V + \lambda\mathcal{L}_R,
\end{equation}
where $\lambda$ is a hyperparameter balancing the importance between meta-verified instruction learning and reflection learning.

By jointly optimizing these objectives during SFT, we train Tool-MVR to master all three System 2 capabilities: (1) reliable tool planning through high-quality API and query verification, (2) accurate tool invocation through verified trajectories, and (3) effective tool reflection through feedback-based error correction.

\section{Experiments}
\subsection{Benchmark and Evaluation Metrics}
\label{sec:benandval}

\subsubsection{StableToolBench}
\label{sec:stabletoolbench} %

We evaluate model performance on StableToolBench~\cite{guo-etal-2024-stabletoolbench}, an enhanced and more stable test set derived from ToolBench~\cite{qintoolllm}. It contains 765 carefully curated tasks that systematically evaluate different aspects of tool learning capabilities. We examine six test scenarios: G1-Ins., G1-Tool, G1-Cat., G2-Ins., G2-Cat., and G3-Ins. These scenarios combine task complexity (\textbf{G1}: single-tool instructions, \textbf{G2}: intra-category multi-tool instructions, \textbf{G3}: cross-category multi-tool instructions) with generalization levels (\textbf{Ins.}: unseen instructions, \textbf{Tool}: unseen tools from seen categories, \textbf{Cat.}: unseen tool categories). Performance is measured using two key metrics: \textbf{Pass Rate} and \textbf{Win Rate}, evaluating task completion success and reasoning quality respectively.

\subsubsection{RefineToolBench}
\label{refinetoolbench}

While existing benchmarks focus primarily on basic tool usage capabilities, they lack systematic assessment of error recovery - a crucial aspect of System 2 reasoning. To address this gap, we introduce RefineToolBench, the first benchmark specifically designed to evaluate tool reflection capabilities. To ensure comprehensive coverage of error scenarios, the benchmark includes both API-level errors in single-tool scenarios ($I1$, $I2$) and complex reasoning failures in multi-tool interactions ($I3$), with carefully designed test cases that require models to recover from errors through reflection. The final dataset contains 344 $I1$ cases (testing parameter and API name errors with available APIs), 388 $I2$ cases (evaluating error handling with simulator-required APIs), and 178 $I3$ cases (assessing complex tool selection and interaction errors). We evaluate using two key metrics: \textbf{Error Recognition Rate} (\textbf{ERR}) measuring error identification capability, and \textbf{Error Correction Rate} (\textbf{ECR}) assessing successful error resolution.

The construction process for RefineToolBench varies by scenario type. For single-tool scenarios (I1 and I2), we use GPT-4 to generate query-action pairs with deliberate API-level errors, ensuring coverage of various error types. For multi-tool scenarios (I3), we first obtain complete reasoning trajectories from StableToolBench using our Stage 1 multi-agent system, then use GPT-4 to introduce errors in tool selection and parameter content at randomly selected steps. All cases undergo rigorous filtering to ensure solvability and meaningful feedback availability, resulting in a comprehensive benchmark for evaluating models' error handling and reflection capabilities.

Comprehensive and detailed evaluation protocols for both benchmarks are provided in Appendix~\cref{sec:evaluation}.

\begin{table*}[t]
\centering
\caption{Pass Rate (\%) comparison in StableToolBench. Bold values indicate the highest scores, \underline{underlined} values indicate the second highest. L and Q represent using LLaMA-3.1-8B-Instruct and Qwen-2.5-7B-Instruct as backbone, respectively.}
\label{tab:main_results}
\small
\begin{tabular}{p{1.2cm}|p{3cm}|cccccc|>{\centering\arraybackslash}p{1cm}p{1cm}}
\toprule
\textbf{Type} & \textbf{Model} & \textbf{G1-Ins.} & \textbf{G1-Tool} & \textbf{G1-Cat.} & \textbf{G2-Ins.} & \textbf{G2-Cat.} & \textbf{G3-Ins.} & \textbf{Average} \\
\midrule
\multirow{2}{*}{Closed} 
& GPT-3.5 & 49.1 & 49.4 & 54.2 & 29.2 & 26.6 & 50.8 & 43.9 \\
& GPT-4 & 74.2 & 72.8 & 60.8 & \underline{63.2} & 66.1 & 75.4 & 68.5 \\
\midrule
\multirow{2}{*}{Base} 
& LLaMA-3.1-8B-Instruct & 53.4 & 53.2 & 54.2 & 42.5 & 48.4 & 49.2 & 50.8 \\
& Qwen-2.5-7B-Instruct & 57.7 & 57.0 & 58.2 & 46.2 & 51.6 & 54.1 & 54.8 \\
\midrule
\multirow{2}{*}{SFT}
& ToolLLM(L) & 60.1 & 60.1 & 67.3 & 50.0 & 46.8 & 70.5 & 58.8 \\
& ToolLLM(Q) & 63.2 & 60.1 & 67.3 & 51.9 & 57.3 & 50.8 & 59.9 \\
\midrule
\multirow{1}{*}{DPO}
& TP-LLaMA(L) & 54.0 & 65.2 & 78.4 & 59.4 & 65.3 & 60.7 & 64.3 \\
\midrule
\multirow{2}{*}{PPO}
& StepTool(L) & 63.8 & 62.7 & 66.7 & 60.4 & 62.9 & 54.1 & 62.7 \\
& StepTool(Q) & 68.7 & 72.8 & 67.3 & 58.5 & 65.3 & 70.5 & 67.5 \\
\midrule
\multirow{6}{*}{\textbf{Ours}} 
& Tool-MVR(L) & \underline{79.1} & \underline{82.9} & 79.1 & \textbf{79.2} & 79.0 & \textbf{86.9} & \underline{80.5} \\
& \;w/o Stage 2(L) & 76.1 & 78.5 & \underline{80.4} & \textbf{79.2} & \underline{80.6} & 78.7 & 78.8 \\
& \;w/o Stage 1\&2(L) & 60.7 & 65.2 & 60.8 & 57.5 & 49.2 & 70.5 & 60.1 \\
& Tool-MVR(Q) & \textbf{83.4} & \textbf{86.7} & \textbf{83.0} & \textbf{79.2} & \textbf{85.5} & \underline{83.6} & \textbf{83.8} \\
& \;w/o Stage 2(Q) & 76.1 & 75.3 & 74.5 & \textbf{79.2} & 74.2 & 82.0 & 76.2 \\
& \;w/o Stage 1\&2(Q) & 46.6 & 43.7 & 54.2 & 47.2 & 47.6 & 41.0 & 47.3 \\
\bottomrule
\end{tabular}
\end{table*}

\subsection{Baselines and Settings}
We compare our Tool-MVR with the following baselines: \textbf{GPT-3.5}~\cite{openai2022chatgpt}, a widely-adopted closed-source LLM that demonstrates excellent performance in tool utilization. \textbf{GPT-4}~\cite{achiam2023gpt4}, the current state-of-the-art closed-source model, exhibits exceptional capabilities in complex reasoning and tool invocation. In the open-source domain, we select two powerful models: \textbf{LLaMA-3.1-8B-Instruct}~\cite{dubey2024llama} and \textbf{Qwen-2.5-7B-Instruct}~\cite{yang2024qwen2}. Using these as backbone models, \textbf{ToolLLM}~\cite{qintoolllm} was developed through supervised fine-tuning on the ToolBench dataset. Subsequently, \textbf{TP-LLaMA}~\cite{chen2024advancing} represents a strong DPO-optimized baseline that enhances inference trajectories by leveraging preference data extracted from decision trees. \textbf{StepTool}~\cite{yu2024steptool} introduces an effective PPO framework that pioneers step-grained reinforcement learning with reward shaping to improve tool learning capabilities. Note that all baselines here are combined with DFSDT for inference.

\subsection{Implementation Details}
\label{Sec:implementation}
We adopt LLaMA-3.1-8B-instruct and Qwen-2.5-7B-instruct as our Tool-MVR backbone and develop our training pipeline based on the LLaMA-Factory~\cite{zheng2024llamafactory} framework. For training loss, we set $\lambda=1$ and maintain a 10:1 ratio between ToolBench-V and ToolBench-R to balance three System 2 reasoning capabilities. The models are optimized using full-parameter fine-tuning with a learning rate of 2e-5 and warmup ratio of 0.04. We leverage DeepSpeed ZeRO-3~\cite{rajbhandari2020zero} optimization and Flash Attention~\cite{dao2023flashattention2} for training acceleration. All experiments are conducted on a machine equipped with 8 NVIDIA A800 GPUs (80GB memory each).

For dataset construction in our ToolBench-V and ToolBench-R, we employ a combination of \enquote{gpt-4-turbo}, \enquote{gpt-4o}, and \enquote{gpt-4} models. In our baseline comparisons, GPT-4 refers to \enquote{gpt-4-turbo}, while GPT-3.5 refers to \enquote{gpt-3.5-turbo}. All evaluation metrics are computed using \enquote{gpt-4-turbo} with temperature set to 0.

\subsection{Data Quality Comparison}
\label{data_quality_analyse}
As discussed in Section~\ref{sec:3.2}, ToolBench suffers from severe quality issues across two dimensions: query quality and API-call quality, both of which are evaluated using three key metrics, following the evaluation protocol in Quality Matters~\cite{iskander2024quality}. To validate the effectiveness of our MAMV pipeline in addressing these limitations, we conducted a comprehensive quality assessment comparing ToolBench-V against the ToolBench dataset, as shown in Table~\ref{tab:metrics}. 
The results demonstrate substantial improvements with our MAMV approach: for user queries, MAMV increases the overall quality score from 52.7\% to 98.8\%, validating our QueryVerifyAgent's effectiveness in filtering out problematic queries. Similarly, for API-call trajectories, our method improves the overall accuracy from 25.6\% to 81.3\% through APICallAgent's comprehensive verification mechanism. For instance, the Execution Checker effectively eliminates redundant API calls and trial-and-error attempts, raising the trajectory minimality score from 54.9\% to 81.9\%. Through these systematic improvements, MAMV successfully addresses the critical limitations in existing datasets: poor query quality, hallucinated API usage, and establishes a strong foundation for tool learning."

\subsection{Main Results on StableToolBench}
We evaluate Tool-MVR against various baselines through three aspects. First, we conduct comprehensive performance comparisons using Pass Rate and Win Rate metrics on StableToolBench, with results presented in Tables \ref{tab:main_results} and \ref{tab:winrate_results}. Second, we perform ablation studies to analyze the contribution of each component in our framework. Third, we provide detailed case studies in \cref{sec:case1} (see Tables~\ref{tab:correct_sequence},~\ref{tab:incorrect_sequence}) to intuitively illustrate the differences between Tool-MVR and ToolLLM in tool planning and invocation capabilities.

\begin{table}[t]
\centering
\caption{Win Rate (\%) comparison in StableToolBench.}
\label{tab:winrate_results}
\small
\begin{tabular}{p{2cm}|p{1.2cm}|ccc|c} 
\toprule
\textbf{Model} & \textbf{Reference} & \textbf{G1} & \textbf{G2} & \textbf{G3} & \textbf{Average} \\
\midrule
GPT-4 & GPT-3.5 & 77.8 & 83.5 & 77.0 & 79.5 \\
ToolLLM(L) & GPT-3.5 & 73.4 & 77.8 & 78.7 & 75.2 \\
ToolLLM(Q) & GPT-3.5 & 72.6 & 78.3 & 60.7 & 73.3 \\
TP-LLaMA(L) & GPT-3.5 & 75.5 & 80.9 & 78.7 & 77.4 \\
StepTool(L) & GPT-3.5 & 75.1 & 80.0 & 70.5 & 76.2 \\
StepTool(Q) & GPT-3.5 & 77.4 & 83.0 & 77.0 & 79.1 \\
\midrule
Tool-MVR(L) & GPT-3.5 & 84.8 & 88.7 & 86.9 & 86.1 \\
Tool-MVR(Q) & GPT-3.5 & \textbf{86.3} & \textbf{89.1} & \textbf{90.2} & \textbf{87.5} \\
\midrule
Tool-MVR(L) & GPT-4 & 75.5 & 78.7 & 78.7 & 76.7 \\
Tool-MVR(L) & ToolLLM & 80.4 & 83.0 & 80.3 & 81.2 \\
Tool-MVR(L) & StepTool & 78.1 & 80.4 & 78.7 & 78.8 \\
\bottomrule
\end{tabular}
\end{table}

\subsubsection{Pass Rate Analysis}
As shown in Tables~\ref{tab:main_results}, we first evaluate models' performance using Pass Rate, which measures their ability to successfully complete tool learning tasks. Our Tool-MVR models, implemented on both LLaMA-3.1-8B (Tool-MVR (L)) and Qwen-2.5-7B (Tool-MVR (Q)), achieve substantial improvements over existing baselines. Tool-MVR (Q) attains an average Pass Rate of 83.8\%, surpassing GPT-4 (by 15.3\%), GPT-3.5 (by 39.9\%), and ToolLLM (Q) (by 23.9\%). Similarly, Tool-MVR (L) achieves 80.5\%, exceeding all LLaMA-based baselines and GPT-4 by 12\%. The superior performance of Tool-MVR can be attributed to two key innovations. First, MAMV's systematic verification ensures high-quality instruction data, which directly improves tool planning and invocation capabilities. Second, EXPLORE's reflection learning mechanism enables models to effectively adapt to and recover from errors.
Beyond performance improvements, our experimental results also demonstrate Tool-MVR's advantages in several key aspects: 1) Generalization Performance: Tool-MVR (Q) achieves its highest Pass Rate of 86.7\% on G1-Tool scenarios with unseen tools from seen categories.
2) Less is More: Tool-MVR achieves these results with only 15,390 training examples, compared to ToolLLM's 73,423, demonstrating that high-quality instruction data matters more than data quantity.
3) Data Quality Impact: While recent approaches like TP-LLaMA and StepTool focus on advanced optimization techniques (DPO and PPO), their modest gains over SFT baselines stem from ToolBench's low-quality data significantly limiting models' basic capabilities. This highlights that improving instruction data quality for SFT is fundamental for enhancing tool learning capabilities.

\subsubsection{Win Rate Analysis}
To further evaluate the quality of tool reasoning beyond simple task completion, we analyze models' Win Rate performance. As shown in Table~\ref{tab:winrate_results}, Tool-MVR (Q) and Tool-MVR (L) demonstrate superior capabilities with Win Rates of 87.5\% and 86.1\% against GPT-3.5 respectively. Notably, Tool-MVR (L) maintains strong performance even against GPT-4 (76.7\%), ToolLLM (81.2\%), and StepTool (78.8\%). These results confirm that our systematic approach to developing tool capabilities - combining high-quality instruction data with reflection learning - not only improves task completion rates but also significantly enhances the quality and completeness of intermediate reasoning processes characteristic of System 2 Reasoning.

\subsubsection{Ablation Study}
To analyze how different components contribute to System 2 reasoning capabilities, we conduct ablation experiments with three configurations. As shown in \cref{tab:main_results}, (1) Tool-MVR (L/Q) represents our complete framework incorporating both meta-verification and reflection learning. (2) \enquote{w/o Stage 2 (L/Q)}, which uses only ToolBench-V without reflection learning, and (3) \enquote{w/o Stage 1\&2 (L/Q)}, which employs the base model with only APICallAgent inference.
Results demonstrate the complementary nature of our two-stage approach. Specifically, (1) the significant performance improvement of \enquote{w/o Stage 2 (Q)} models over ToolLLM (Q) (76.2\% vs. 59.9\% ) validates MAMV's effectiveness in developing tool planning and invocation abilities; (2) the further enhancement achieved by the complete Tool-MVR (Q) (83.8\%) highlights EXPLORE's crucial role in strengthening tool reflection capabilities; (3) the substantially lower performance of \enquote{w/o Stage 1\&2 (L)} (47.3\%) confirms that both components are essential for comprehensive System 2 reasoning in tool learning.

\subsection{Tool Reflection Results on RefineToolBench}
\label{sec:4.5}
To systematically evaluate models' tool reflection capabilities, we conduct experiments on RefineToolBench across three error scenarios: single API errors with available APIs ($I1$), single API errors requiring simulators ($I2$), and multi-API interaction errors ($I3$). As shown in Table~\ref{tab:error_performance}, we assess both Error Recognition Rate (ERR) and Error Correction Rate (ECR) to measure models' ability to identify and rectify mistakes. Additionally, we provide detailed case studies in the appendix (see \cref{sec:case2} and Tables~\ref{tab:correct_sequence1},~\ref{tab:wrong_sequence}) to qualitatively analyze the differences in reflection capabilities.

\subsubsection{Limitations of Existing Approaches}
Our analysis reveals a critical limitation in existing Tool-Augmented LLMs: their poor tool reflection capabilities. ToolLLM(L) and ToolLLM(Q) achieve only 7.8\% and 9.1\% Error Correction Rate (ECR) respectively. These models repeatedly make the same mistakes without adaptation when encountering errors. Notably, base models (LLaMA-3.1-8B-Instruct and Qwen-2.5-7B-Instruct) show better error handling (29.7\% and 35.4\% ECR), indicating that static imitation learning on ToolBench actually impairs models' inherent error recovery capabilities by focusing solely on successful examples while ignoring valuable learning signals from failures.

\begin{table}[t]
\centering
\small
\setlength{\tabcolsep}{4pt}
\caption{Error Recognition Rate (\%) and Correction Rate (\%) across different error scenarios in RefineToolBench. Note: LLaMA-3.1-8B and Qwen-2.5-7B refer to instruction versions.}
\label{tab:error_performance}
\begin{tabular}{l|cccc|cccc}
\toprule
\multirow{2}{*}{\textbf{Method}} & \multicolumn{4}{c|}{\textbf{Error Recognition}} & \multicolumn{4}{c}{\textbf{Error Correction}} \\
& \textbf{I1} & \textbf{I2} & \textbf{I3} & \textbf{Avg.} & \textbf{I1} & \textbf{I2} & \textbf{I3} & \textbf{Avg.} \\
\midrule
GPT-3.5 & 42.2 & 45.9 & 66.3 & 48.5 & 39.0 & 44.6 & 60.1 & 45.5 \\
GPT-4 & 50.0 & 45.4 & 81.5 & 54.2 & 48.0 & 44.1 & 72.5 & 51.1 \\
LLaMA-3.1-8B & 30.8 & 37.1 & 24.7 & 32.3 & 28.8 & 34.8 & 20.2 & 29.7 \\
Qwen-2.5-7B & 46.2 & 48.2 & 28.1 & 43.5 & 36.3 & 41.0 & 21.3 & 35.4 \\
ToolLLM (L) & 6.1 & 4.1 & 34.8 & 10.9 & 4.9 & 2.8 & 24.2 & 7.8 \\
ToolLLM (Q) & 5.2 & 4.9 & 36.5 & 11.2 & 5.2 & 4.9 & 25.8 & 9.1 \\
\midrule
Tool-MVR (L) & \underline{59.9} & \underline{57.7} & \underline{88.8} & \underline{64.6} & \underline{51.2} & \underline{52.3} & \textbf{81.5} & \underline{57.6} \\
\;w/o  Stage 2 (L) & 51.2 & 50.0 & 78.1 & 55.9 & 43.3 & 45.9 & 68.0 & 49.2 \\
Tool-MVR (Q) & \textbf{63.7} & \textbf{60.3} & \textbf{90.4} & \textbf{67.5} & \textbf{53.2} & \textbf{54.4} & \underline{79.8} & \textbf{58.9} \\
\;w/o  Stage 2 (Q) & 56.1 & 53.9 & 82.6 & 60.3 & 46.2 & 46.6 & 66.9 & 50.4 \\
\bottomrule
\end{tabular}
\end{table}

\subsubsection{Performance Improvements}
Tool-MVR demonstrates substantial improvements in reflection capabilities across all scenarios. Tool-MVR(Q) achieves an average ERR of 67.5\% and ECR of 58.9\%, significantly outperforming GPT-4 (54.2\% ERR, 51.1\% ECR), while Tool-MVR(L) attains 64.6\% ERR and 57.6\% ECR. These results validate the effectiveness of our dynamic \enquote{Error $\rightarrow$ Reflection $\rightarrow$ Correction} learning paradigm, which enables models to systematically learn from and adapt to execution feedback.

\subsubsection{Ablation Analysis}
Ablation studies further validate the importance of reflection learning. As shown in Table~\ref{tab:error_performance}, removing Stage 2 (EXPLORE) leads to significant performance drops: ERR decreases from 67.5\% to 60.3\%, and ECR drops from 58.9\% to 50.4\%, demonstrating EXPLORE's crucial role in error handling. The benefits of reflection capabilities extend beyond RefineToolBench: on StableToolBench, the pass rate improves from 76.2\% to 83.8\% through enhanced tool reflection abilities from Stage 2 (EXPLORE).

\subsubsection{Performance Analysis Across Different Scenarios}
Tool-MVR shows varying performance across error scenarios: lower ECR on single-tool cases (53.2\% and 54.4\% for $I1$ and $I2$) compared to multi-tool scenarios (79.8\% for $I3$). This pattern aligns with the training data distribution, as $I3$ cases from StableToolBench contain familiar multi-tool tasks, while single-tool scenarios present entirely novel challenges. Notably, despite this domain gap, Tool-MVR(Q) maintains strong generalization ability, achieving substantial improvements over GPT-4 (48.0\%, 44.1\%, 72.5\%) across all scenarios.

\subsection{Efficiency Analysis}

\begin{figure}[t]
\centering
\includegraphics[width=.8\linewidth]{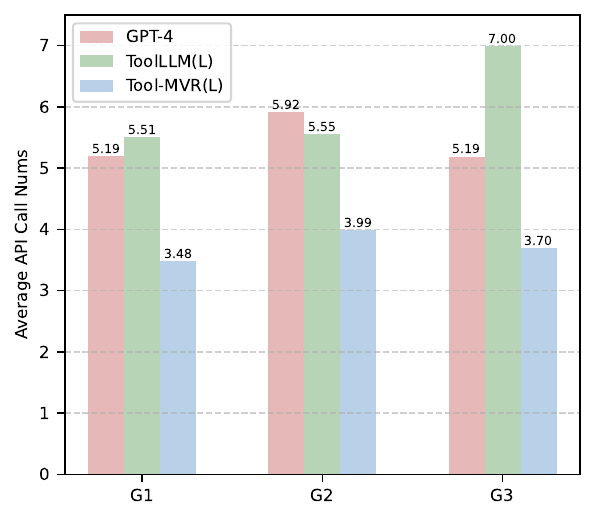}
\caption{Average number of API calls required by different models across test scenarios in StableToolBench.}
\label{fig: steps}
\end{figure}

As shown in Figure~\ref{fig: steps}, Tool-MVR achieves superior efficiency in StableToolBench, requiring only 3.48-3.99 API calls per task across scenarios. This represents a 37.4\% reduction compared to ToolLLM and 31.4\% compared to GPT-4, while maintaining state-of-the-art performance. This significant reduction in API calls substantially decreases user interaction latency and reduces operational costs in commercial applications.
Our MAMV pipeline achieves this efficiency improvement by significantly enhancing training data quality, reaching a minimality score of 81.9\% compared to ToolBench's 54.9\%, as shown in Table~\ref{tab:metrics}. High-quality training examples with minimal and accurate API calls enable Tool-MVR to develop efficient tool usage patterns, eliminating redundant operations during inference. This combination of improved efficiency and effectiveness demonstrates Tool-MVR's enhanced capabilities in deliberate tool planning and precise tool invocation.

\section{Conclusions}
In this paper, we introduced Tool-MVR, which successfully enhanced open-source LLMs' System 2 reasoning capabilities and achieved state-of-the-art performance in tool learning through two key innovations. First, we developed MAMV to establish high-quality instruction data through meta-verification, addressing the critical limitation of instruction dataset quality and establishing strong foundations for deliberate tool planning and accurate tool invocation. Second, we proposed EXPLORE to enable systematic error correction through reflection learning, addressing a critical gap in existing models' reflection capabilities that resulted from static imitation learning. Our experiments demonstrated that Tool-MVR achieved state-of-the-art performance on StableToolBench, surpassing both open-source and closed-source baselines while requiring significantly fewer API calls, and exhibited strong generalization capabilities across unseen tools and scenarios. Additionally, on RefineToolBench - the first benchmark for evaluating tool reflection capabilities - Tool-MVR achieved the highest error recognition and correction rates, demonstrating unprecedented adaptive tool reflection capabilities.

\bibliographystyle{ACM-Reference-Format}
\bibliography{sample-sigconf}

\appendix

\section{Experimental Evaluation Details}
\label{sec:evaluation}

We employ four comprehensive metrics to evaluate model performance, with standardized evaluation protocols implemented through GPT-4:

\textbf{Pass Rate (PR)} evaluates the overall task completion success by examining both final answers and execution chains (\cref{tab:passrate_status_prompt}). A Pass requires sufficient query resolution with successful API calls and verifiable information, while a Fail indicates execution errors or invalid solutions. Cases lacking sufficient validation data or complete reasoning processes are marked as Unsure:
\begin{equation}
    PR = \frac{1}{N}\sum_{i=1}^{N} s_i, \text{ where } s_i = \begin{cases} 
    1.0 & \text{if Pass} \\
    0.5 & \text{if Unsure} \\
    0.0 & \text{if Fail}
    \end{cases}
\end{equation}

\textbf{Win Rate (WR)} conducts pairwise comparisons between solutions using a comprehensive 100-point scoring system (\cref{tab:winrate_status_prompt}). The evaluation considers answer quality (60\%) - assessing completeness of required information (20pts), accuracy with API responses (20pts), presentation clarity (10pts), and error documentation (10pts) - and execution efficiency (40\%) - evaluating appropriate tool selection and usage (15pts), logical progression (15pts), and strategic optimization (10pts):
\begin{equation}
    WR = \frac{\sum_{i=1}^{M} \mathbb{1}(C(A_i, B_i) = 0)}{M}
\end{equation}
where $C(A_i, B_i)$ returns 0 if $A_i$ is superior to $B_i$, with ties resolved in favor of higher answer quality scores.

Following the error handling evaluation protocol (\cref{tab:error_eval_prompt}), we assess two complementary aspects:

\textbf{Error Recognition Rate (ERR)} measures the model's ability to explicitly identify API errors or invalid responses. A Pass requires clear error acknowledgment, while proceeding without recognizing errors results in a Fail:
\begin{equation}
    ERR = \frac{\sum_{i=1}^{K} \mathbb{1}(r_i = \text{Pass})}{K}
\end{equation}

\textbf{Error Correction Rate (ECR)} evaluates both error recognition and successful resolution, requiring valid subsequent API calls and proper error handling. This metric captures the complete error recovery process:
\begin{equation}
    ECR = \frac{\sum_{i=1}^{K} \mathbb{1}(r_i = \text{Pass} \land c_i = \text{Pass})}{K}
\end{equation}

Here, $N$ denotes the total number of tasks, $M$ the number of comparison pairs, and $K$ the number of error cases. All evaluations require complete execution chains and API responses for comprehensive assessment.

\subsection{Case Study 1: Tool Planning and Invocation}
\label{sec:case1}

To better understand Tool-MVR's enhanced System 2 reasoning capabilities, we present a detailed case study comparing Tool-MVR with ToolLLM. Given a query requesting both a comprehensive language list for an event and specific airplane details for an aviation presentation (shown in Table~\ref{tab:correct_sequence}), the two approaches demonstrate stark differences in their reasoning capabilities.

Tool-MVR demonstrates sophisticated System 2 reasoning through systematic planning and precise execution (Table~\ref{tab:correct_sequence}). In Step 1, it begins with a comprehensive analysis of the task, explicitly recognizing the need to obtain valid airplane IDs before querying specific details. Following this analysis, Tool-MVR executes an efficient three-step solution: first retrieving the language list (Step 1), then fetching the airplane list to identify suitable IDs (Step 2), and finally obtaining detailed information about the Boeing 737-800 using ID 1 (Step 3). Each step builds logically on previous results, with no redundant API calls, culminating in a complete solution that addresses both the language list and airplane details in its final response (Step 4).

In stark contrast, ToolLLM's approach reveals several critical limitations (Table~\ref{tab:incorrect_sequence}). Its tool planning exhibits characteristics of System 1's quick, intuitive thinking - after retrieving the language list (Step 1), it immediately attempts to query airplane details with an arbitrary ID 12345 (Step 2) without proper planning. When this fails, ToolLLM does successfully retrieve the airplane list (Step 3) containing valid IDs including the Boeing 737-800 (ID: 1). However, instead of utilizing this information, it demonstrates severely flawed tool invocation patterns:

\begin{itemize}
    \item Redundant API Calls: Unnecessarily retrieves the same language list three times (Steps 1, 4, and 6)
    \item Repeated Errors: Attempts the same invalid airplane ID 12345 twice (Steps 2 and 5), despite previous failure
    \item Ignored Information: Fails to use the valid airplane IDs obtained in Step 3
\end{itemize}

Most critically, ToolLLM's final response (Step 7) exhibits classic hallucination behavior - providing only the language list while completely omitting airplane information, despite having access to valid airplane data in Step 3. This mirrors the broader issue of hallucinated trajectories in ToolBench leading to incomplete solutions.

Tool-MVR addresses these limitations through our two-stage approach. MAMV's systematic verification ensures high-quality instruction data in ToolBench-V, enabling models to learn proper tool planning and invocation patterns, as demonstrated by the logical progression and efficient API usage in Table~\ref{tab:correct_sequence}. Additionally, EXPLORE's reflection-based learning replaces blind backtracking with deliberate error analysis and strategy adjustment, preventing the type of repetitive errors seen in ToolLLM's approach.

The stark contrast between Tool-MVR's systematic three-step solution and ToolLLM's chaotic seven-step process validates our approach's effectiveness in developing true System 2 reasoning capabilities. While ToolLLM exhibits trial-and-error behavior characteristic of System 1 thinking - making repeated mistakes without learning from them - Tool-MVR demonstrates deliberate planning, precise execution, and complete solution generation through systematic verification and reflection-based learning.

\subsection{Case Study 2: Tool Reflection}
\label{sec:case2}

To demonstrate how Tool-MVR handles real-world API challenges through EXPLORE's reflection learning, we present a detailed comparison of successful and failed approaches to an order retrieval task (shown in Tables~\ref{tab:correct_sequence1} and~\ref{tab:wrong_sequence}). This case study is particularly relevant as it deals with two common challenges in production environments: incomplete API documentation and evolving parameter specifications.

The task involves interacting with two poorly documented APIs: \texttt{get\_user\_orders\_for\_onboarding\_project\_v3()} for retrieving order history and \texttt{get\_order\_for\_onboarding\_project\_v3(is\_id: str)} for detailed order information. The documentation presents real-world challenges: missing parameter specifications for the first API and an outdated parameter name (\texttt{is\_id} instead of \texttt{order\_id}) for the second - reflecting how API implementations often evolve faster than their documentation.

Tool-MVR demonstrates sophisticated System 2 reasoning through its dynamic interaction with these imperfect APIs (Table~\ref{tab:correct_sequence1}). Its systematic error correction process includes:
\begin{itemize}
    \item \textbf{Parameter Discovery:} When encountering "Input parameters missing" error in Step 1, it analyzes the feedback and discovers the need for user identification, successfully adding it in Step 2
    \item \textbf{Type Adaptation:} Upon receiving "Invalid API Request" in Step 3, it identifies the parameter type mismatch and attempts integer conversion in Step 4
    \item \textbf{Parameter Evolution:} Despite documentation showing \texttt{is\_id}, it discovers and implements the correct \texttt{order\_id} parameter in Step 5 through feedback analysis
\end{itemize}

In contrast, the failed approach (Table~\ref{tab:wrong_sequence}) exhibits classic System 1 behavior with multiple critical failures:
\begin{itemize}
    \item \textbf{Feedback Ignorance:} Despite receiving explicit error message "Missing required parameters" in Step 3, it continues with empty parameter calls
    \item \textbf{Strategy Stagnation:} Repeats identical wrong API calls four times without modification
    \item \textbf{Learning Failure:} Misses key information in error messages, such as "you need to provide the user ID" in Step 3
    \item \textbf{Premature Abandonment:} Gives up after Step 4 without attempting parameter adjustments or exploring alternative approaches
\end{itemize}

This stark contrast validates EXPLORE's effectiveness in developing true System 2 reasoning for tool reflection. While the failed approach demonstrates ``fast thinking'' - making quick, repetitive attempts without reflection - Tool-MVR exhibits deliberate ``slow thinking'' through structured error analysis and strategic adjustment. This capability, enabled by EXPLORE's systematic ``Error → Reflection → Correction'' cycles, transforms API errors into learning opportunities, effectively handling real-world scenarios such as ambiguous user queries, incomplete API documentation, and evolving API specifications across different versions.

\section{Implementation Details of Tool-MVR}
\label{sec:implementation}

\subsection{APIOptAgent Details}
\label{sec:apiopt}

APIOptAgent enhances API documentation through three components: Validator, Simulator, and DocRefiner. The Validator first generates diverse API calls following the example generation prompt shown in Table~\ref{tab:example_prompt} to obtain real execution observations. These observations serve as empirical evidence for API validation and documentation refinement. For instance, when testing a car data API, the Validator systematically explores different parameter combinations (e.g., varying page numbers, makes, and models) to understand the API's actual behavior and limitations.

For invalid or unstable APIs, the Simulator creates standardized responses following the system prompt presented in Table~\ref{tab:system_prompt}, which ensures consistent error handling and response structure. The DocRefiner then leverages both successful and failed API calls to improve documentation quality using the refinement prompt described in Table~\ref{tab:refine_prompt}. Following OpenAI's function calling format, the refined documentation is structured as a JSON object with standardized fields including ``type'', ``function'', ``name'', ``description'', and ``parameters'', ensuring compatibility with existing LLM tool-use frameworks.

The effectiveness of this empirical refinement process is demonstrated in Table~\ref{tab:api_doc_comparison}, where vague documentation is transformed based on observed API behaviors. The original description ``This is the subfunction for tool car\_data'' is enhanced with concrete pagination and filtering capabilities discovered through actual API interactions. Parameter descriptions are similarly refined based on observed valid inputs and error cases, ensuring documentation accurately reflects real API functionality. This observation-driven approach significantly improves documentation quality by grounding specifications in empirical evidence rather than theoretical assumptions, while maintaining strict adherence to OpenAI's standardized API documentation format.

\subsection{QueryVerifyAgent Details}
\label{sec:queryverify}

QueryVerifyAgent evaluates query solvability and quality through a systematic verification process following the template shown in Table~\ref{tab:query_verify_prompt}. The agent determines query solvability based on four key rules: (1) queries containing invalid or nonsensical information are marked as unsolvable, (2) queries missing essential information required by the APIs are unsolvable, (3) queries that cannot be mapped to available tools are unsolvable, and (4) queries with sufficient valid information that match available API capabilities are marked as solvable. Additionally, each query is assigned a quality score from 1 to 10 based on five criteria: solvability (ease of solving using tools), semantic clarity (query clarity and grammar), information completeness (necessary detail inclusion), reasoning difficulty (complexity of required reasoning), and tool compatibility (alignment with tool capabilities). For example, the query ``I need to keep track of the transaction volume on the BSC testnet and receive webhook notifications for new transactions'' would be marked as unsolvable despite having clear intent - while the transaction volume part is solvable, the webhook notification request lacks required transaction IDs for API setup, making the complete query unsolvable.

\begin{figure*}[h]
\centering
\includegraphics[width=\linewidth]{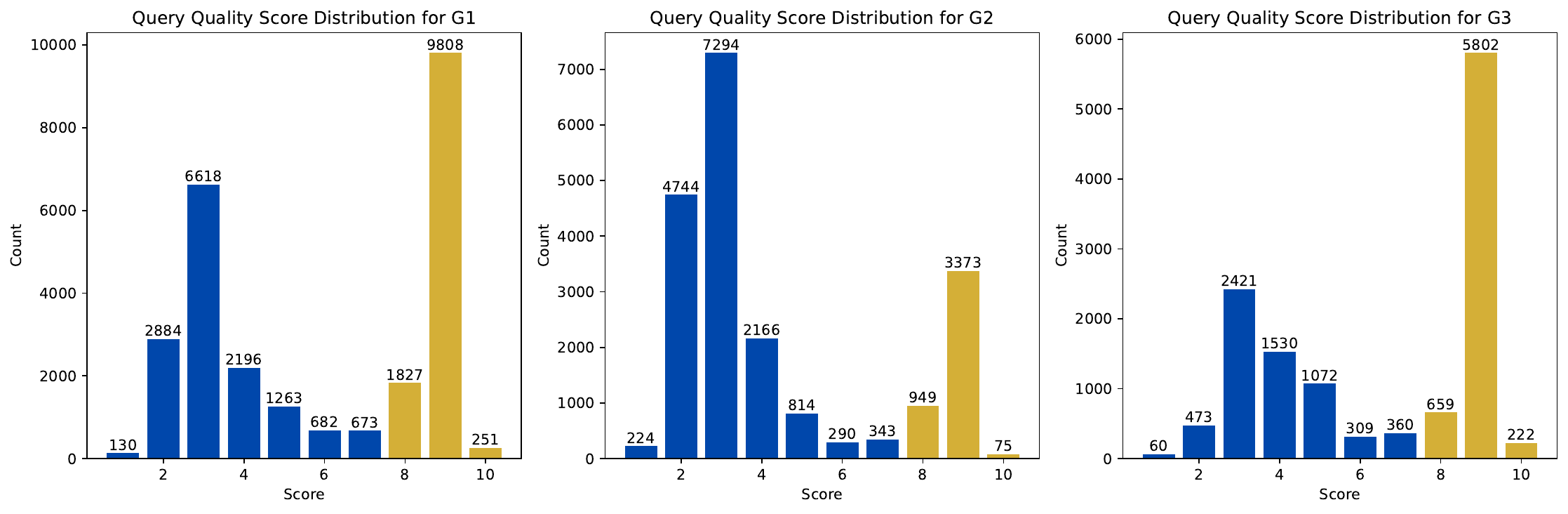}
\caption{Distribution of query quality scores across different groups (G1, G2, G3) in ToolBench. The yellow bars indicate high-quality queries (scores 8-10) that are retained in our filtered dataset.}
\label{fig: quality}
\end{figure*}

Our analysis reveals substantial variations in query solvability across groups. G3 demonstrates the highest solvability rate at 50.7\% of 14,361 queries, while G1 achieves 47.0\% of 27,059 queries. G2 shows notably lower solvability at 23.4\% of 20,727 queries, indicating potential mismatches between queries and available tools. To ensure high-quality instruction data, we retain only queries with quality scores between 8 and 10, as highlighted in yellow in Figure~\ref{fig: quality}. This filtering process results in 11,776 queries from G1 (43.5\% retention), 4,347 from G2 (21.0\% retention), and 6,664 from G3 (46.4\% retention).

The retained high-quality queries demonstrate clear intent, complete information, and strong alignment with available tools. For instance, queries like ``I want to analyze the options data for the stock symbol `AAPL'. Can you fetch the options data for this stock?'' provide all necessary parameters and match well with available API capabilities. This rigorous quality control ensures that our instruction dataset effectively develops models' tool planning capabilities, enabling them to accurately understand user intent and systematically plan appropriate API selections.

\subsection{APICallAgent Details}
\label{sec:apicall}

APICallAgent constructs high-quality API call trajectories through a two-stage process: trajectory construction and verification. The trajectory construction stage follows the prompt template shown in Table~\ref{tab:apicall_prompt}, which guides the agent to break down complex queries into sub-problems and solve them iteratively through API calls. Unlike traditional JSON-based function calling formats, we adopt Python function calls to enable more sophisticated operations including complex data processing and transformation, conditional branching based on API responses, cascading API calls with intermediate result handling, and adaptive  parameter adjustment based on previous call results.

For each sub-problem, the agent must provide its reasoning in a \texttt{<thought>} tag before making API calls in an \texttt{<execute>} block. This structured format ensures transparent decision-making process, clear connection between reasoning and actions, traceable problem-solving steps, and flexible adaptation to API feedback.

The verification stage implements a comprehensive triple-verification mechanism using the template in Table~\ref{tab:answer_verify_prompt}. The Format Checker validates whether the trajectory contains all required components, particularly focusing on the presence of a \texttt{<final\_answer>} tag. Trajectories lacking a final answer or containing incomplete responses are filtered out, ensuring that each retained trajectory represents a complete solution attempt.

The Semantic Checker evaluates whether the answer comprehensively resolves the user's query. It examines information completeness by verifying all aspects of the query are addressed, logical consistency by ensuring the reasoning aligns with API responses, and answer validity by confirming the solution's correctness based on API outputs. Trajectories that only partially solve the problem or contain incorrect solutions are eliminated, maintaining high solution quality.

The Execution Checker assesses the accuracy and necessity of each step in the trajectory. It validates that each API call meaningfully contributes to the final solution and eliminates trajectories containing redundant or trial-and-error steps. This ensures trajectory optimality by retaining only those solutions where each step plays a crucial role in problem resolution. The checker also verifies proper error handling and parameter selection throughout the trajectory.

The verification process assigns each trajectory an answer status (``Pass'', ``Fail'', or ``Unsure'') and validates step necessity. Only trajectories that pass all verification stages are retained, ensuring the final dataset contains only complete, accurate, and efficient solutions. This rigorous filtering eliminates common issues such as incomplete answers, partial solutions, redundant API calls, and trial-and-error attempts.

Through this comprehensive construction and verification pipeline, APICallAgent successfully generates reliable and efficient API call trajectories that serve as high-quality training examples for developing robust tool invocation capabilities. The combination of Python-based flexibility and rigorous verification ensures that models learn to generate accurate parameter values and execute APIs in the correct format, leading to precise and efficient tool invocation during inference.

\subsection{Reflection Details from EXPLORE}
\label{sec:reflection}

To systematically develop tool reflection capabilities characteristic of System 2 reasoning, we implement a structured reflection mechanism. This mechanism enables models to dynamically process execution feedback and adjust their reasoning strategies through three key steps:
(1) \textbf{Error Analysis}: identifying specific issues in API selection or parameter generation from execution feedback.
(2) \textbf{Reflection Generation}: developing correction strategies through deliberate analysis of API documentation and previous attempts.
(3) \textbf{Action Planning}: formulating precise execution plans with corrected API calls.

\begin{table*}[h]
\centering
\renewcommand{\arraystretch}{1.3}
\caption{Pass Rate evaluation prompt template.}
\label{tab:passrate_status_prompt}
\small
\begin{tabular}{|p{\textwidth}|}
\hline
\textbf{Pass Rate Evaluation Prompt} \\ \hline
You are an assistant responsible for evaluating whether an LLM agent's response should be counted as Pass, Fail, or Unsure in passrate calculations. Your evaluation must consider both the final answer and the complete execution chain. \\
\textbf{Status Determination Rules:} \\
\textbf{Pass:} Answer sufficiently solves query; Execution chain shows successful API calls; Initial errors were corrected; Information verifiable through API responses \\
\textbf{Fail:} API observations show execution errors; Answer contradicts evidence; Information incorrect/invalid; Solution misses core requirements \\
\textbf{Unsure:} Cannot verify authenticity; Insufficient validation data; Need complete reasoning process; No useful information despite attempts \\
\textbf{Output Format:} \\
\begin{minipage}{\textwidth}
\begin{verbatim}
{"content": "Evaluation reasoning", "answer_status": "Pass/Fail/Unsure"}
\end{verbatim}
\end{minipage} \\
\textbf{Required Input:} Original query; Final answer; Complete execution chain with API responses \\ [0.5em]
\hline
\end{tabular}
\end{table*}
\begin{table*}[h]
\centering
\renewcommand{\arraystretch}{1.3}
\caption{Win Rate evaluation prompt template.}
\label{tab:winrate_status_prompt}
\small
\begin{tabular}{|p{\textwidth}|}
\hline
\textbf{Win Rate Evaluation Prompt Template} \\ \hline
You are an assistant responsible for comparing two answers (Answer\_0 and Answer\_1) to determine which solution is superior. Your evaluation must follow a weighted scoring system to select the better answer. \\
\textbf{Evaluation Criteria (100 points):} \\
\textbf{1. Answer Quality (60 points):} 
Completeness (20pts): Contains all required information, fully addresses query requirements;
Accuracy (20pts): Information correctness, alignment with API responses;
Clarity (10pts): Well-structured presentation, easy to understand;
Error Handling (10pts): Clear explanation of failures, proper error documentation \\
\textbf{2. Execution Efficiency (40 points):} 
Tool Usage (15pts): Failed API calls (-2pts each), redundant calls (-1pt each), appropriate tool selection;
Execution Path (15pts): Logical progression, minimal steps, efficient goal achievement;
Strategy (10pts): Tool selection planning, response adaptation, resource optimization \\
\textbf{Output Format:} \\
\begin{minipage}{\textwidth}
\begin{verbatim}
{"content": "Comparative analysis", "better_answer_index": "0 or 1"}
\end{verbatim}
\end{minipage} \\
\textbf{Required Input:} Original query text; Answer\_0 and Answer\_1 with complete execution chains \\
\textbf{Tiebreaker Rule:} In case of equal scores, prefer the answer with higher quality score \\ [0.5em]
\hline
\end{tabular}
\end{table*}

\begin{table*}[h]
\centering
\renewcommand{\arraystretch}{1.3}
\caption{Error Recognition Rate and Error Correction Rate evaluation prompt template.}
\label{tab:error_eval_prompt}
\small
\begin{tabular}{|p{\textwidth}|}
\hline
\textbf{Error Recognition and Error Correction Rate Evaluation Prompt Template} \\ \hline
You are an assistant responsible for evaluating a model's ability to recognize and correct errors in API interactions. Your evaluation must determine both error recognition status and error correction status. \\
\textbf{Evaluation Rules:} \\
\textbf{Error Recognition (Pass/Fail):} 
Model must explicitly acknowledge API errors or invalid responses;
Fail if model proceeds without recognizing the error;
Only Pass when clear indication of error awareness is shown \\
\textbf{Error Correction (Pass/Fail):} 
Model must successfully resolve the identified error;
Subsequent API calls must yield valid results;
Fail if error persists or workaround is unsuccessful \\
\textbf{Combined Status Rules:} 
Both Pass: Must recognize AND successfully correct error;
Both Fail: No error recognition leads to automatic correction failure;
Mixed Case: Can Pass recognition but Fail correction if error identified but not resolved \\
\textbf{Output Format:} \\
\begin{minipage}{\textwidth}
\begin{verbatim}
{
    "content": "Evaluation reasoning",
    "error_recognition": "Pass/Fail",
    "error_correction": "Pass/Fail"
}
\end{verbatim}
\end{minipage} \\
\textbf{Required Input:} \\
- Initial messages (query and API context) \\
- Wrong action messages (failed API calls/responses) \\
- After messages (subsequent handling) \\
\textbf{Constraints:} Only Pass/Fail values allowed; No other status values permitted \\ [0.5em]
\hline
\end{tabular}
\end{table*}

\begin{table*}[h]
\centering
\renewcommand{\arraystretch}{1.3}
\caption{Case study 1: Tool-MVR's correct solution - demonstrating systematic tool planning and precise API invocation.}
\label{tab:correct_sequence}
\small
\begin{tabular}{|p{\textwidth}|}
\hline
\textbf{Input Details} \\ \hline
\textbf{Query:} I'm organizing a language learning event and I need a list of languages to include in the program. Can you provide me with a comprehensive list of languages? Additionally, I would like to know the details of a specific airplane for a presentation on aviation history. \\ [1em]

\textbf{Available APIs:} \\
\texttt{search\_roman\_god\_by\_name\_for\_roman\_gods\_by\_pizza\_api(god: str)} - Search god by name for details and Greek equivalents \\
\texttt{random\_roman\_god\_for\_roman\_gods\_by\_pizza\_api()} - Get random Roman god info \\
\texttt{get\_list\_of\_languages\_for\_businessmate()} - Get supported languages with codes \\
\texttt{endpoint1\_for\_chattydata()} - Get HTML formatted static response \\
\texttt{single\_airplane\_for\_airplanesdb(is\_id: int)} - Get specific airplane details by ID \\
\texttt{all\_airplanes\_for\_airplanesdb()} - Get all airplanes data \\
\texttt{airplanes\_ordered\_by\_ascending\_for\_airplanesdb(ordering: str)} - Get airplanes ordered by field \\ [1em]
\hline

\textbf{Step 1: Initial Planning and Language List} \\ \hline

\textbf{Thought:} To solve this task, I need to perform the following steps: 1. Retrieve a list of languages supported by the `businessmate' tool for the language learning event. 2. Retrieve detailed information about a specific airplane using the `single\_airplane\_for\_airplanesdb' API. To do this, I need a valid airplane ID. Since the task doesn't provide an ID for the airplane, I will first fetch the list of all airplanes using `all\_airplanes\_for\_airplanesdb' to identify a specific airplane to study further. \\

\textbf{Action:} get\_list\_of\_languages\_for\_businessmate() \\

\textbf{Observation:} \{"error": "", "response":\{"success": true, "message": "success", "results": [\{"text": "AFRIKAANS", "value": "af"\}, \{"text": "ALBANIAN", "value": "sq"\}, \{"text": "AMHARIC", "value": "am"\}, \ldots]\}\} \\ [1em]
\hline

\textbf{Step 2: Getting Airplane Options} \\ \hline

\textbf{Thought:} I have successfully retrieved a comprehensive list of languages. Next, I need to fetch the list of all airplanes using the `all\_airplanes\_for\_airplanesdb' API to identify a specific airplane ID for further detailed retrieval. This will allow me to select an appropriate aircraft for the aviation history presentation. \\

\textbf{Action:} all\_airplanes\_for\_airplanesdb() \\

\textbf{Observation:} \{"error": "", "response":\{"id":1,"plane":"Boeing 737-800","brand":"Boeing","passenger\_capacity":189,\ldots,"id":2,"plane":"A220 100",\ldots\}\} \\ [1em]
\hline

\textbf{Step 3: Getting Specific Airplane Details} \\ \hline

\textbf{Thought:} I have retrieved the list of all airplanes. For the presentation on aviation history, I can use details about the ``Boeing 737-800" (plane ID: 1) due to its widespread use and significance in commercial aviation. Next, I will retrieve detailed information about this specific airplane using its ID. \\

\textbf{Action:} single\_airplane\_for\_airplanesdb(is\_id=1) \\

\textbf{Observation:} \{"error": "", "response":\{"id": 1, "plane": "Boeing 737-800", "brand": "Boeing", "passenger\_capacity": 189, "priceUSD": 84400000.0, "fuel\_capacity\_litres": 26020.0, "max\_takeoff\_weight\_kg": 79000.0, \ldots\}\} \\ [1em]
\hline

\textbf{Step 4: Final Response} \\ \hline

\textbf{Thought:} I have successfully retrieved both required pieces of information: 1. A comprehensive list of languages for the language learning event, which includes multiple options across different language families and regions. 2. Detailed specifications of the Boeing 737-800, including its technical specifications, performance capabilities, and commercial details. Now I can provide a complete response that addresses both aspects of the original query. \\

\textbf{Action:} given\_final\_answer \\

\textbf{Final Answer:} For your language learning event, you can include these languages: Afrikaans, Albanian, Amharic, Arabic, Armenian, Azerbaijani, and more. For the aviation presentation, key details of the Boeing 737-800: \textbullet{} Passenger Capacity: 189 \textbullet{} Price: \$84,400,000 \textbullet{} Fuel Capacity: 26,020 liters \textbullet{} Range: 5,400 km \textbullet{} Engine: CFMI CFM56-7B24/26 \\ [0.5em]
\hline
\end{tabular}
\end{table*}

\begin{table*}[h]
\centering
\renewcommand{\arraystretch}{1.3}
\caption{Case study 1: ToolLLM's incorrect solution - showing flawed tool planning and redundant, erroneous API calls.}
\label{tab:incorrect_sequence}
\small
\begin{tabular}{|p{\textwidth}|}

\hline
\textbf{Input Details} \\ \hline
\textit{Query and Available APIs are identical to Table~\ref{tab:correct_sequence}} \\ [0.3em]
\hline
\textbf{Step 1: Get Languages (First Attempt)} \\ \hline

\textbf{Thought:} To solve this task, I need to perform the following steps: 1. Retrieve a list of languages supported by the `businessmate' tool for the language learning event. 2. Retrieve detailed information about a specific airplane using the `single\_airplane\_for\_airplanesdb' API. Let me start with getting the languages... \\

\textbf{Action:} get\_list\_of\_languages\_for\_businessmate() \\

\textbf{Observation:} \{"error": "", "response": \{"success": true, "results": [\{"text": "AFRIKAANS", "value": "af"\}, \{"text": "ALBANIAN", "value": "sq"\}, \{"text": "AMHARIC", "value": "am"\}, \{"text": "ARABIC", "value": "ar"\}, \{"text": "ARMENIAN", "value": "hy"\},...]\}\} \\ [1em]
\hline

\textbf{Step 2: Wrong Airplane ID (First Attempt)} \\ \hline

\textbf{Thought:} I have successfully retrieved a comprehensive list of languages. Next, I need to fetch the details of a specific airplane. I will use the `single\_airplane\_for\_airplanesdb' API with ID 12345 to get airplane information for the presentation... \\

\textbf{Action:} single\_airplane\_for\_airplanesdb(is\_id=12345) \\

\textbf{Observation:} \{"error": "", "response": "[]"\} \\ [1em]
\hline

\textbf{Step 3: Get All Airplanes (But Not Used)} \\ \hline

\textbf{Thought:} Based on the previous actions and the current state, it seems that the attempt to retrieve the details of a specific airplane using the `single\_airplane\_for\_airplanesdb' function with ID 12345 did not return any results. To proceed, I will try a different approach and use the `all\_airplanes\_for\_airplanesdb' function... \\

\textbf{Action:} all\_airplanes\_for\_airplanesdb() \\

\textbf{Observation:} \{"error": "", "response": [\{"id": 1, "plane": "Boeing 737-800", "brand": "Boeing", "passenger\_capacity": 189, "priceUSD": 84400000.0, "fuel\_capacity\_litres": 26020.0\}, \{"id": 2, "plane": "A220 100", "brand": "Airbus",...\}]\} \\ [1em]
\hline

\textbf{Step 4: Get Languages Again (Unnecessary Repetition 1)} \\ \hline

\textbf{Thought:} The current API call is to retrieve a list of supported languages for the `businessmate' tool. This API call is made to fulfill the user's request for a comprehensive list of languages. By calling this API, I can provide the user with the necessary information to include in their language learning event program... \\

\textbf{Action:} get\_list\_of\_languages\_for\_businessmate() \\

\textbf{Observation:} \{"error": "", "response": \{"success": true, "results": [\{"text": "AFRIKAANS", "value": "af"\}, \{"text": "ALBANIAN", "value": "sq"\}, \{"text": "AMHARIC", "value": "am"\}, \{"text": "ARABIC", "value": "ar"\},...]\}\} \\ [1em]
\hline

\textbf{Step 5: Wrong Airplane ID Again (Repeated Mistake)} \\ \hline

\textbf{Thought:} I have successfully retrieved a comprehensive list of languages. Next, I will proceed to retrieve the details of a specific airplane using the `single\_airplane\_for\_airplanesdb' function. \textbf{I will provide the ID 12345} to gather the required information for the user's presentation on aviation history... \\

\textbf{Action:} single\_airplane\_for\_airplanesdb(\textbf{is\_id=12345}) \\

\textbf{Observation:} \{"error": "", "response": "[]"\} \\ [1em]
\hline

\textbf{Step 6: Get Languages Yet Again (Unnecessary Repetition 2)} \\ \hline

\textbf{Thought:} Based on the previous action, it seems that the attempt to retrieve details of a specific airplane using the `single\_airplane\_for\_airplanesdb' function failed. Since the previous action did not provide any useful information, I will try a different approach. To fulfill the user's request for a comprehensive list of languages, I will use the `get\_list\_of\_languages\_for\_businessmate' function again... \\

\textbf{Action:} get\_list\_of\_languages\_for\_businessmate() \\

\textbf{Observation:} \{"error": "", "response": \{"success": true, "results": [\{"text": "AFRIKAANS", "value": "af"\}, \{"text": "ALBANIAN", "value": "sq"\}, \{"text": "AMHARIC", "value": "am"\}, \{"text": "ARABIC", "value": "ar"\}, \{"text": "ARMENIAN", "value": "hy"\},...]\}\} \\ [1em]
\hline

\textbf{Step 7: Final Response} \\ \hline

\textbf{Thought:} The function `single\_airplane\_for\_airplanesdb' was called to retrieve the details of a specific airplane. However, the function output indicates that the request was unsuccessful and an empty response was returned. Since I couldn't retrieve the details of the specific airplane, I will provide the list of languages obtained from the previous action as the final answer for the language learning event... \\

\textbf{Action:} given\_final\_answer \\

\textbf{Final Answer:} Here is a comprehensive list of languages: AFRIKAANS... \textbf{[Incomplete response - missing airplane details]} \\ [0.5em]
\hline
\end{tabular}
\end{table*}

\begin{table*}[h]
\centering
\renewcommand{\arraystretch}{1.3}
\caption{Case study 2: Tool-MVR's correct solution - demonstrating tool reflection in API calling and strategy adaptation based on API feedback.}
\label{tab:correct_sequence1}
\small
\begin{tabular}{|p{\textwidth}|}
\hline
\textbf{Input Details} \\ \hline
\textbf{Query:} Can you fetch the details of my recent orders? I would like to know the products I ordered, the order status, and the delivery date. \\ [1em]
\textbf{Available APIs:} \\
\texttt{\textbf{get\_user\_orders\_for\_onboarding\_project\_v3()}} - Get user's order history \\
\texttt{get\_order\_for\_onboarding\_project\_v3(is\_id: str)} - Get detailed information for a specific order \\
... (other available APIs omitted) \\ [1em]
\hline

\textbf{Step 1: Initial Order History Attempt} \\ \hline

\textbf{Thought:} To fetch the details of recent orders, I will use the `get\_user\_orders\_for\_onboarding\_project\_v3' API. This API should provide information regarding the products ordered, order status, and delivery date. \\

\textbf{Action:} get\_user\_orders\_for\_onboarding\_project\_v3() \\

\textbf{Observation:} \{"error": "Input parameters missing; please provide the user ID or name to retrieve orders.", "response": ""\} \\ [1em]
\hline

\textbf{Step 2: Adding User Identification} \\ \hline

\textbf{Thought:} The previous API call returned an error because I didn't provide the required user identifier. \textbf{I need to adjust my strategy and pass the user ID or name as a parameter to retrieve the orders. Since the documentation doesn't specify which parameter to use, I will try with `user\_id' first.} If that doesn't work, I will try `name'. \\

\textbf{Action:} get\_user\_orders\_for\_onboarding\_project\_v3(user\_id="my\_user\_id") \\

\textbf{Observation:} \{"error": "", "response": \{"user\_id": "my\_user\_id", "orders": [\{"order\_id": "1234", ...\}, \{"order\_id": "1235", ...\}], ...\}\} \\ [1em]
\hline

\textbf{Step 3: First Attempt at Detailed Order Info} \\ \hline

\textbf{Thought:} The API call was successful, and I retrieved the user's orders. However, the response lacks information on the order status and delivery date. I will need to use the `get\_order\_for\_onboarding\_project\_v3' API with the `order\_id' to fetch more detailed information, including the order status and delivery date. \\

\textbf{Action:} get\_order\_for\_onboarding\_project\_v3(is\_id="1234") \\

\textbf{Observation:} \{\textbf{"error": "Invalid API Request", "response": "The parameters provided do not match with any of our API requirements}..."\} \\ [1em]
\hline

\textbf{Step 4: Adjusting Parameter Type} \\ \hline

\textbf{Thought:} The error message indicates that there is a mismatch in the parameters provided. \textbf{It seems that the parameter should be an integer, not a string. I will adjust the strategy and try passing the parameter as an integer.} \\

\textbf{Action:} get\_order\_for\_onboarding\_project\_v3(is\_id=1234) \\

\textbf{Observation:} \{"error": "The 'Get Order' API has insufficient information in its description...", "response": ""\} \\ [1em]
\hline

\textbf{Step 5: Final Parameter Adjustment} \\ \hline

\textbf{Thought:} \textbf{The error message suggests that the `is\_id' parameter is unclear or not documented for this API. It might be an issue with the parameter type or name. Given the situation, I will try to use the `order\_id' as a last resort since it's the primary key for the order.} I will call the `get\_order\_for\_onboarding\_project\_v3' API again but this time using `order\_id' as the parameter. \\

\textbf{Action:} get\_order\_for\_onboarding\_project\_v3(\textbf{order\_id=1234}) \\

\textbf{Observation:} \{"error": "", "response": \{"order\_id": 1234, "customer": "John Doe", "items": [\{"product\_name": "Apple iPhone 12", ...\}], "status": "Shipped"\}\} \\ [1em]
\hline

\textbf{Step 6: Final Response} \\ \hline

\textbf{Thought:} I successfully retrieved the detailed information for the order using the `get\_order\_for\_onboarding\_project\_v3` API. The information includes the customer name, items ordered, their quantities and prices, the total price, and the order status. Although the delivery date is not available, the order status is mentioned as "Shipped." Given the details I have, I will now summarize the information for the recent orders placed by the user. \\
\textbf{Action:} given\_final\_answer \\
\textbf{Final Answer:} Your recent orders are as follows:
1. Order ID: 1234, Product: Apple iPhone 12, Quantity: 1, Price: \$699, Status: Shipped
2. Order ID: 1235, Product: Product B, Quantity: 1, Price: \$24.99, Status: Not available

The delivery date for the above orders is not available. \\ [0.5em]
\hline
\end{tabular}
\end{table*}

\begin{table*}[h]
\centering
\renewcommand{\arraystretch}{1.3}
\caption{Case study 2: ToolLLM's failed solution - showing repetitive incorrect API calling without adjustment leading to abandonment.}
\label{tab:wrong_sequence}
\small
\begin{tabular}{|p{\textwidth}|}
\hline
\textbf{Input Details} \\ \hline
\textbf{Query:} Can you fetch the details of my recent orders? I would like to know the products I ordered, the order status, and the delivery date. \\ [1em]

\textbf{Available APIs:} \\
\texttt{get\_user\_orders\_for\_onboarding\_project\_v3()} - Get user's order history \\
\texttt{get\_order\_for\_onboarding\_project\_v3(is\_id: str)} - Get detailed information for a specific order \\
... (other available APIs omitted) \\ [1em]
\hline

\textbf{Step 1: Initial Attempt} \\ \hline

\textbf{Thought:} Based on the user's request to fetch the details of their recent orders, I need to call the `get\_user\_orders\_for\_onboarding\_project\_v3' function. This function will retrieve the user's orders from the onboarding project. By calling this function with the appropriate arguments, I can obtain the necessary information about the products ordered, the order status, and the delivery date. This will help me provide the user with the requested details and fulfill their query. \\

\textbf{Action:} get\_user\_orders\_for\_onboarding\_project\_v3() \\

\textbf{Observation:} \{\textbf{"error": "Incomplete input parameters"}, "response": \{\}\} \\ [1em]
\hline

\textbf{Step 2: Repeated Empty Attempt} \\ \hline

\textbf{Thought:} I apologize, but it seems that there is an error with the input parameters. P\textbf{lease provide the necessary information to fetch your recent orders.} \\

\textbf{Action:} \textbf{get\_user\_orders\_for\_onboarding\_project\_v3()} \\

\textbf{Observation:} \{"error": "Input is incomplete or unclear.", "response": ""\} \\ [1em]
\hline

\textbf{Step 3: Another Empty Attempt} \\ \hline

\textbf{Thought:} I apologize, but it seems that there is still an error with the input parameters. \textbf{Please provide the necessary information to fetch your recent orders.} \\

\textbf{Action:} \textbf{get\_user\_orders\_for\_onboarding\_project\_v3()} \\

\textbf{Observation:} \{"error": "Missing required parameters", "response": "To retrieve the user's orders, you need to provide the user ID. Please refer to the API documentation for necessary parameters."\} \\ [1em]
\hline

\textbf{Step 4: Final Failed Attempt} \\ \hline

\textbf{Thought:} Based on the previous action and the current state, it seems that the input parameters for fetching the user's orders are missing. \textbf{To resolve this, I will provide the necessary user ID as an argument when calling the `get\_user\_orders\_for\_onboarding\_project\_v3' function.} \\

\textbf{Action:} \textbf{get\_user\_orders\_for\_onboarding\_project\_v3()} \\

\textbf{Observation:} \{"error": "Invalid input: No parameters provided.", "response": \{\}\} \\ [1em]
\hline
\textbf{Step 5: Giving Up} \\ \hline
\textbf{Thought:} It seems that the function `get\_user\_orders\_for\_onboarding\_project\_v3' requires input parameters to retrieve the user orders. \textbf{Since I don't have the necessary input parameters, I won't be able to fetch the details of your recent orders. I will need to restart the task and try again.} \\
\textbf{Action:} \textbf{given\_up\_and\_restart} \\
\textbf{Final Answer:} given\_up\\ [1em]
\hline
\end{tabular}
\end{table*}

\begin{table*}[h]
\centering
\renewcommand{\arraystretch}{1.5}
\caption{Prompt template for generating API call examples.}
\label{tab:example_prompt}
\small
\begin{tabular}{|p{\textwidth}|}
\hline
\textbf{Example Generation Prompt} \\  \hline
Please generate a few example function calls based on the following API documentation, using keyword arguments to fully understand the API functionality. The output should be in Python format with printed results in code blocks. Maximum 3 example calls, and results must be printed. 
\\ 
\textbf{Note:} No need to import API package names, just call the API functions directly. If the API has no parameters, just call the API function directly. \\ 
\textbf{Example Format:} \\ 
\begin{minipage}{\textwidth}
\begin{verbatim}
print("Example 1:")
print(ticket_info_query(destination="Beijing", travel_mode="Train"))

print("Example 2:")
print(ticket_info_query(departure="Shanghai", destination="Beijing", travel_mode="Plane"))

print("Example 3:")
print(ticket_info_query(departure="Guangzhou", destination="Shenzhen", travel_mode="Bus"))
\end{verbatim}
\end{minipage} \\ 
\textbf{Input:} API name, API documentation \\ [0.8em]
\hline
\end{tabular}
\end{table*}

\begin{table*}[h]
\centering
\renewcommand{\arraystretch}{1.5}
\caption{Prompt template for refining API documentation.}
\label{tab:refine_prompt}
\small
\begin{tabular}{|p{\textwidth}|}
\hline
\textbf{API Documentation Refinement Prompt} \\  \hline
You are a expert in API documentation. Now we have a few API call results. Please refine the API description based on the API call results to better describe the API functionality. Note you can only change the API description and parameters description, you can't change the API name and parameters name and you can't add or remove parameters.
\\ 
\textbf{Response Format:} \\ 
\begin{minipage}{\textwidth}
\begin{verbatim}
# For invalid/no-data APIs:
{"is_api_valid": false}

# For valid APIs (no changes):
{"is_api_valid": true}

# For APIs needing refinement:
{
    "is_api_valid": true,
    "refine_api": {
        "type": "function",
        "function": {"name": "...", "description": "...",
        "parameters": {"properties": {...}, "required": [...]}}}
}
\end{verbatim}
\end{minipage} \\ 
\textbf{Input:} API name, Original doc, API call, Observation \\ [0.8em]
\hline
\end{tabular}
\end{table*}

\begin{table*}[h]
\centering
\renewcommand{\arraystretch}{1.5}
\caption{System prompt template for API Simulation.}
\label{tab:system_prompt}
\small
\begin{tabular}{|p{\textwidth}|}
\hline
\textbf{API Simulation Prompt} \\  \hline
Imagine you are an API Server operating within a specialized tool, which contains a collection of distinct APIs. Your task is to process the given input parameters and construct a meaningful, relevant, and structured JSON response based on these inputs and we provide API descriptions in the API documentation. Analyze the input carefully to understand its intended purpose, and generate the corresponding output data. Your response must follow the structure below:
\\ 
\begin{minipage}{\textwidth}
\begin{verbatim}
{
    "error": "",
    "response": "<Your_Response>"
}
\end{verbatim}
\end{minipage} \\ 
\textbf{Example:} \\ 
\begin{minipage}{\textwidth}
\begin{verbatim}
# API Doc
GetFinalExamScores: Retrieves student's exam scores
Input: {"student_id": "12345"}
Output: {
    "error": "",
    "response": {
        "student_id": "12345",
        "scores": [{"subject": "Math", "score": 95}]
    }
}
\end{verbatim}
\end{minipage} \\ 
\textbf{Input:} API Documentation, API Input, API Output \\ [0.5em]
\hline
\end{tabular}
\end{table*}

\begin{table*}[h]
\centering
\renewcommand{\arraystretch}{1.3}
\caption{Comparison of original and refined API documentation.}
\label{tab:api_doc_comparison}
\small
\begin{tabular}{|p{0.3\textwidth}|p{0.6\textwidth}|}
\hline
\textbf{Original API Documentation} & \textbf{Refined API Documentation} \\ \hline
\textbf{Name:} cars\_for\_car\_data & \textbf{Name:} cars\_for\_car\_data \\ \hline
\textbf{Description:} This is the subfunction for tool ``car\_data'', you can use this tool. The description of this function is: ``Retrieve and filter lists of cars'' & 
\textbf{Description:} ``Retrieve and filter lists of cars based on parameters such as page, limit, make, year, model, and type. Use this function to fetch paginated car data, ensuring correct parameter configuration to avoid empty responses.'' \\ \hline
\multicolumn{2}{|l|}{\textbf{Parameters and Description:}} \\ \hline
\textbf{page} (required): ``'' & \textbf{page} (required): ``The page number for pagination. Starts from 0.'' \\ \hline
\textbf{limit} (required): ``'' & \textbf{limit} (required): ``The maximum number of car records to retrieve per page.'' \\ \hline
\textbf{make} (optional): ``'' & \textbf{make} (optional): ``The manufacturer or brand of the car (e.g., Toyota, Tesla).'' \\ \hline
\textbf{year} (optional): ``'' & \textbf{year} (optional): ``The manufacturing year of the car.'' \\ \hline
\textbf{model} (optional): ``'' & \textbf{model} (optional): ``The specific model of the car (e.g., Corolla, Escape).'' \\ \hline
\textbf{type} (optional): ``'' & \textbf{type} (optional): ``The type or body style of the car (e.g., Sedan, SUV, Electric).'' \\ \hline
\end{tabular}
\end{table*}

\begin{table*}[h]
\centering
\renewcommand{\arraystretch}{1.3}
\caption{Query verification prompt template.}
\label{tab:query_verify_prompt}
\small
\begin{tabular}{|p{\textwidth}|}
\hline
\textbf{Query Verification Prompt} \\  \hline
You are an assistant responsible for evaluating whether a given user query can be solved using the provided APIs and tools. Your evaluation must consider both the completeness of the query's information, the capabilities of the available APIs, and the overall quality of the query. \\
\textbf{Rules:} \\
1. \textbf{Invalid Information:} If the query contains invalid or nonsensical information, return ``Unsolvable''. \\
2. \textbf{Missing Information:} If the query lacks essential information required to solve it, return ``Unsolvable''. \\
3. \textbf{Incomplete or Ambiguous Tools:} If unable to determine how to solve the query with provided tools, return ``Unsolvable''. \\
4. \textbf{Solvable Query:} If the query provides sufficient valid information and available APIs/tools can solve it, return ``Solvable''. \\
\textbf{Quality Scoring (1-10):} \\
- Solvability: Ease of solving using provided tools \\
- Semantic Clarity: Query clarity and grammar \\
- Information Completeness: Necessary detail inclusion \\
- Reasoning Difficulty: Complexity of required reasoning \\
- Tool Compatibility: Alignment with tool capabilities \\
\textbf{Output Format:} \\
1. Explanation of decision \\
2. JSON response with decision and quality score \\
\textbf{Example:} \\
Query: ``Can you fetch the flight data for the company AZU on June 15th, 2022?'' \\
Tools: Flight data API with company and date parameters \\
Output: \{``decision'': ``Solvable'', ``quality\_score'': 7\} \\
\textbf{Input Required:} Query text, Available tools list \\ [0.5em]
\hline
\end{tabular}
\end{table*}

\begin{table*}[h]
\centering
\renewcommand{\arraystretch}{1.3}
\caption{Prompt template for API call trajectory construction.}
\label{tab:apicall_prompt}
\small
\begin{tabular}{|p{\textwidth}|}
\hline
\textbf{API Call Trajectory Construction Prompt} \\  \hline
You are an AutoGPT, capable of utilizing numerous APIs to complete the given task. You will use an interactive programming environment and retrieve the latest knowledge through the provided APIs. The APIs are real and have been defined and imported—do not define new APIs. The API documentation may be limited or brief, but you must try to use them. When calling an API, carefully read the documentation, use appropriate parameters, and ensure correct values. If there is no parameter in documentation, use the API directly. If the API returns nothing, retry and adjust your strategy. Each sub-question requires an API call to solve. \\

\textbf{Problem-Solving Process:} \\
Break the problem into sub-steps and solve iteratively. In each iteration, complete one of: \\

1. \textbf{API Call Required:} \\
   - Provide analysis in \texttt{<thought>} tag: \\
     \texttt{<thought>} I need to get the price of Hilton hotels in Shanghai. \texttt{</thought>} \\
   - Call API with \texttt{<execute>} tag and Python code: \\
     \texttt{<execute>} \\
     \begin{verbatim}
     print(query_hotel_list_by_demand(city="Shanghai", demand="Hilton Hotel"))
     \end{verbatim}
     \texttt{</execute>} \\
   - Ensure logical flow between \texttt{<thought>} and \texttt{<execute>} \\
   - Call only subfunctions' names, not original tool names \\
   - Maximum 2 API calls per iteration \\

2. \textbf{Problem Resolution:} \\
   - Summarize reasoning in \texttt{<thought>} tag \\
   - Conclude with either: \\
     - \texttt{<final\_answer>}: Provide solution \\
     - \texttt{<given\_up>}: Indicate inability to solve \\

\textbf{Core Rules:} \\
Each iteration must include: \\
- One \texttt{<thought>} tag \\
- Followed by either \texttt{<execute>}, \texttt{<final\_answer>}, or \texttt{<given\_up>} \\

\textbf{Input Required:} Query text and API docs \\ [0.5em]
\hline
\end{tabular}
\end{table*}

\begin{table*}[h]
\centering
\renewcommand{\arraystretch}{1.3}
\caption{Answer verification prompt template.}
\label{tab:answer_verify_prompt}
\small
\begin{tabular}{|p{\textwidth}|}
\hline
\textbf{Answer Verification Prompt} \\ \hline
You are an assistant responsible for verifying the quality and correctness of answers by analyzing both the query and its reasoning process. Your evaluation must consider the answer status and the validity of reasoning steps. \\
\textbf{Answer Status Rules:} \\
\textbf{Pass:} If confident the answer sufficiently solves the query or contains successful tool executions \\
\textbf{Fail:} If API errors occur, final answer is invalid, or information contradicts observations \\
\textbf{Unsure:} If unable to determine solution status or API provides no valid information \\
\textbf{Step Validity Analysis:} \\
Evaluate each reasoning step's contribution to final result \\
Check for irrelevant or incorrect attempts \\
Verify logical progression of steps \\
Assess necessity of each API call \\
\textbf{Output Format:} \\
\begin{minipage}{\textwidth}
\begin{verbatim}
{
"content": "Verification explanation",
"answer_status": "Pass/Fail/Unsure",
"all_steps_validity": "yes/no"
}
\end{verbatim}
\end{minipage} \\
\textbf{Example:} \\
Query: ``What's the weather in New York?'' \\
Process: API call successful, temperature retrieved \\
Output: \{``content'': ``Valid weather data obtained'', ``answer\_status'': ``Pass'', ``all\_steps\_validity'': ``yes''\} \\
\textbf{Input Required:} Query text, Reasoning process and answer \\ [0.5em]
\hline
\end{tabular}
\end{table*}

\begin{table*}[h]
\centering
\renewcommand{\arraystretch}{1.3}
\caption{Reflection generation prompt template.}
\label{tab:reflection_prompt}
\small
\begin{tabular}{|p{\textwidth}|}
\hline
\textbf{Reflection Generation Prompt} \\  \hline
Given a query and several initial iterations with the last being incorrect, generate a reflection for self-correction. \\

\textbf{Instructions:} \\
1. \textbf{Context Understanding:} Review API documentation in system prompt containing multiple APIs with documentation. Analyze previous iterations that attempted to solve the query. Examine the incorrect iteration's \texttt{<thought>} and \texttt{<execute>} components. Consider the observation feedback from the failed attempt. \\

2. \textbf{Error Analysis Categories:} \\
\textbf{API Selection Errors:} Using incorrect API names that differ from the correct ones, applying wrong API parameter values leading to incorrect results, attempting to use non-existent APIs not in the provided list. \\
\textbf{Parameter Errors:} Omitting required parameters in API calls, providing invalid input parameters, type mismatches (e.g., using strings for integer parameters), date format errors, parameter values outside valid ranges, logically meaningless parameters despite correct formatting. \\

3. \textbf{Reflection Generation Format:} \\
\texttt{<thought>} \\
Analyze the observation feedback to identify the specific error type and its root cause. For execution errors, examine API call correctness and parameter validity. For successful but meaningless executions, evaluate API choice relevance. Plan step-by-step correction strategy and formulate proper API selection and parameter configuration. \\
\texttt{</thought>} \\
\texttt{<execute>} \\
\begin{verbatim}
print(corrected_api_call(parameter1="value1",parameter2="value2"))
\end{verbatim}
\texttt{</execute>} \\

\textbf{Core Rules:} Must include both \texttt{<thought>} and \texttt{<execute>} components. Reflection should simulate a self-correcting LLM capable of proper API selection and parameter understanding. Analysis must be based on observation feedback rather than prior knowledge of errors. Thought process should be thorough and detailed for training purposes. Error discovery should be active based on feedback, not predetermined. \\

\textbf{Reference:} Correct iteration provided as reference for proper solution approach: \texttt{===right\_iteration===} \\

\textbf{Output Format:} \\
\# Reflection \\
\texttt{<thought>} Detailed error analysis and correction strategy \texttt{</thought>} \\
\texttt{<execute>} \\
\begin{verbatim}
# correct api call
\end{verbatim}
\texttt{</execute>} \\

\textbf{Input Required:} Query text, previous iterations, incorrect iteration with feedback, reference correct iteration \\ [0.5em]
\hline
\end{tabular}
\end{table*}

\end{document}